\documentclass[lettersize,journal]{IEEEtran}

\usepackage{xcolor}
\usepackage{color}

\usepackage{flushend} 

\usepackage{cite}

\usepackage[colorlinks,
linkcolor=blue,
anchorcolor=blue,
citecolor=blue]{hyperref}

\usepackage{amsmath}

\usepackage{algorithmic}

\usepackage[ruled,linesnumbered]{algorithm2e}

\usepackage{array}

\usepackage{stfloats}

\usepackage{graphicx} 
\usepackage{float}  
\usepackage{subfigure} 
\graphicspath{ {./images/} }

\usepackage{enumitem}

\usepackage{booktabs}

\usepackage{bbding}
\usepackage{pifont}
\usepackage{wasysym}
\usepackage{amssymb}

\usepackage{multirow}

\usepackage{xr}

\usepackage{ragged2e}

\begin{document}

\title{Position-Aware Relation Learning for RGB-Thermal Salient Object Detection}

\author{Heng Zhou,~\IEEEmembership{Graduate Student Member,~IEEE,}
	Chunna Tian$^*$,
	Zhenxi Zhang,~\IEEEmembership{Graduate Student Member,~IEEE,}
	\\
	Chengyang Li,~\IEEEmembership{Graduate Student Member,~IEEE,}
	Yuxuan Ding,
	Yongqiang Xie$^*$,
	and
	Zhongbo Li,

	\thanks{Manuscript received September 20, 2022; 
		This research was supported by 
		the National Natural Science Foundation of China under Grant No.62173265.
		\textit{(Corresponding author: Chunna Tian, Yongqiang Xie.)}
	}
	\vspace{-2em}}  

\markboth{IEEE Transactions on Image Processing,~Vol.~XX, No.~X, September~2022}%
{Shell \MakeLowercase{\textit{et al.}}: A Sample Article Using IEEEtran.cls for IEEE Journals}

\IEEEpubid{0000--0000/00\$00.00~\copyright~2022 IEEE}

\maketitle

\begin{abstract}
RGB-Thermal salient object detection (SOD) combines two spectra to segment visually conspicuous regions in images.
Most existing methods use boundary maps to learn the sharp boundary.
These methods ignore the interactions between isolated boundary pixels and other confident pixels, leading to sub-optimal performance.
To address this problem,
we propose a position-aware relation learning network (PRLNet) for RGB-T SOD based on swin transformer.
PRLNet explores the distance and direction relationships between pixels to strengthen intra-class compactness and inter-class separation, generating salient object masks with clear boundaries and homogeneous regions.
Specifically, we develop a novel signed distance map auxiliary module (SDMAM) to improve encoder feature representation, which takes into account the distance relation of different pixels in boundary neighborhoods.
Then, we design a feature refinement approach with directional field (FRDF), which rectifies features of boundary neighborhood by exploiting the features inside salient objects.
FRDF utilizes the directional information between object pixels to effectively enhance the intra-class compactness of salient regions.
In addition, we constitute a pure transformer encoder-decoder network to enhance multispectral feature representation for RGB-T SOD.
Finally, we conduct quantitative and qualitative experiments on three public benchmark datasets.
The results demonstrate that our proposed method outperforms the state-of-the-art methods.
\end{abstract}


\section{Introduction}
\label{sec:Intro}
\IEEEPARstart{S}{alient} object detection (SOD) is to segment the main conspicuous objects in the image at the pixel level by simulating the human visual system.
In applications of image quality assessment~\cite{gu2016saliency,chen2021depth}, image editing~\cite{chen2020rgbd,feng2019attentive}, person re-identification~\cite{zhou2019discriminative} and robotics~\cite{dawson2020provably,zhou2020hierarchical}, 
SOD extracts informative objects in images to help scene analysis and understanding.
Traditional SOD methods mainly use low-level features and certain priors, such as color contrast and background priors, to detect targets~\cite{wang2021salient}.

In recent years, CNN-based SOD methods~\cite{zhang2019rgb,liu2020multi,zhang2021summarize,zhou2022edge} have shown advantages over traditional hand-crafted feature-based methods in terms of model accuracy and generalization.
The application of SOD is also extended from visible light images to multispectral ones~\cite{song2020multi}.
Thermal sensors rely on the thermal radiation of the object to generate images,
which are not easily affected by environmental conditions, such as weather, illumination, \textit{etc}.~\cite{bondi2020birdsai}.
For example, the quality of thermal images is noticeably better than RGB images in low illumination.
RGB-T image pairs have both the radiometric intensity of infrared and the detail information of visible light.
Compared with single RGB images, RGB-T multispectral fusion can generate discriminative and robust saliency features~\cite{hao2019hsme,zhou2022multi}.
Therefore, the RGB-T SOD method achieves a more robust generalization performance in real-world scenes.

\IEEEpubidadjcol

\begin{figure}[t]  
	\centering 
	\includegraphics[scale=0.35]{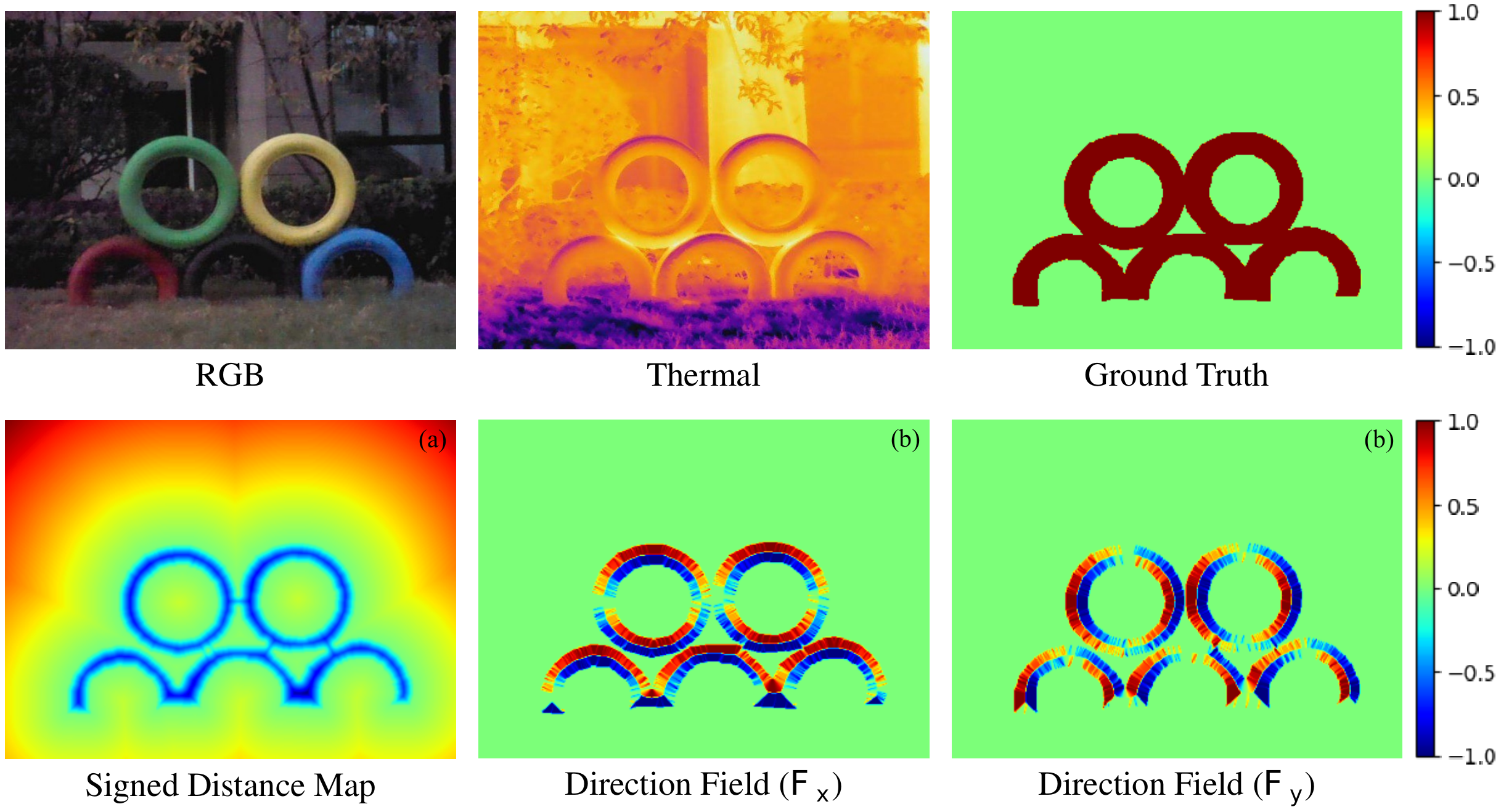}
	\caption{RGB-T SOD with position-aware relation learning.
	\textbf{(a)}~The signed distance map (SDM) calculates the distance from a pixel to the nearest boundary,
	where the sign indicates that the pixel is inside~(+) or outside~(-) the salient object.
	The zero level set is the boundary of the salient object.
	\textbf{(b)}~The directional field ($ \mathcal{F}_{x}, \mathcal{F}_{y} $) of a salient object points from the pixel to its nearest boundary pixel.
}
	\label{Fig.supervise_inf} 
\end{figure}

To obtain accurate salient object masks, many CNN-based models~\cite{qin2019basnet,zhang2022collaborative,zhou2022edge} focus on generating clear contours by learning edge maps.
The consistency between pixels mainly includes intra-class compactness and inter-class separation.
However, these methods ignore the relation learning between boundary pixels and region pixels (target and background regions), resulting in unsatisfactory results.
In this paper, we propose a position-aware relation learning network (PRLNet) to model the distance relations and direction relations between pixels, which enhances the intra-class compactness and inter-class separability of features.

The relative distance information between pixels can effectively alleviate the misprediction of salient pixels~\cite{zhang2021location}.
Inspired by the level set method~\cite{zhang2019resls,cai2021avlsm}, the signed distance map (SDM) models the distance relation between region pixels (target and background regions) and boundary pixels.
As shown in Fig.~\ref{Fig.supervise_inf}, 
SDM provides object boundary interaction information which is the distance between the foreground-background region and the boundary. 
SDM prediction task can be served as an auxiliary task to guide the feature extraction of the encoder.
Different from multi-task learning of SDM in decoder~\cite{farag2013novel,chai2020aerial,lin2021bsda},
we propose the SDM auxiliary module (SDMAM) to enhance the boundary-awareness of the encoder.
SDMAM assists the encoder to learn the relative distance between the region pixels and the boundary, and increases the inter-class separation of the pixels.

Not only the distance relationship, but also the directional relationship between salient pixels is crucial in position-aware relationship learning.
Fig.~\ref{Fig.supervise_inf} shows the visualization of the horizontal and vertical directions of the direction field~\cite{cheng2020learning}. 
As illustrated in Fig.~\ref{Fig.supervise_inf}, the direction field can simply yet efficiently represent the directional information between intra-class pixels and boundary pixels. 
To strengthen the compactness of intra-class pixels, we propose a feature refinement approach with direction field (FRDF) to rectify the output feature maps of the decoder. 
FRDF exploits the direction relationship between saliency pixels to effectively reinforce the compactness of intra-class features.
Meanwhile, we design a novel direction-aware loss function to improve the smoothing loss~\cite{godard2017unsupervised,wang2018occlusion}, which guides the model to generate homogeneous regions and sharp boundaries.

In addition, we take full advantage of transformer~\cite{vaswani2017attention} in modeling long-range contexts to generate cross-spectral robust RGB-T features.
Swin transformer~\cite{liu2021swin} adopts a hierarchical architecture to effectively solve the visual multi-scale problem and reduce the computational complexity,
achieving state-of-the-art (SOTA) performance in semantic segmentation and instance segmentation~\cite{liu2021swin,he2022swin,yu2022soit}.
In our paper, we use swin transformer as the backbone network for RGB-T feature extraction.
At the same time, we design a novel patch separating layer to upsample the encoder features, which build a swin transformer decoder. 
At the same time, the reverse swin transformer is designed to decode the RGB-T patches.
Finally, we propose a position-aware relation learning network (PRLNet) based on pure transformer for RGB-T SOD.

In summary, the main contributions of this paper are as follows.

\begin{itemize}
	\item We propose the novel PRLNet with pure swin transformer to generate salient object masks with clear boundaries and homogeneous regions by learning the distance and direction relationships between different pixels.
	\item Specifically, the SDM auxiliary module is suggested to learn the distance relation of each pixel to the boundary, enhancing boundary-based inter-class (foreground-background) separation.
	
	\item In order to strengthen the intra-class compactness, we design a feature refinement approach with direction field (FRDF) and direction-aware smoothness loss.

	The features close to the boundary are refined by utilizing the internal features of salient objects.

	\item The qualitative and quantitative experimental results on three public benchmark datasets demonstrate that our proposed model outperforms the state-of-the-art models.
	
\end{itemize}

The rest of this paper is organized as follows.
Section~\ref{sec:relatedWorks} overviews the existing methods mainly on RGB and RGB-T SOD and swin transformer.
In Section~\ref{sec:method}, we introduce our proposed position-aware relation learning network for RGB-T SOD.
Extensive experiments and visualization results on the three benchmark datasets are given in Section~\ref{sec:exp}. 
Finally, we conclude our work in Section~\ref{sec:conclusion}.

\section{Related works}
\label{sec:relatedWorks}
In this section, we review the previous SOD methods for RGB and RGB-T images.
Meanwhile, related works about swin transformer are also included in this section.

\subsection{RGB Salient Object Detection}
Recently, most CNN-based SOD methods adopt a fully convolutional network (FCN) structure~\cite{long2015fully,chen2020reverse}.
To improve the accuracy of prediction results, multi-level feature fusion~\cite{zhang2018deep,deng2018r3net,wu2022edn} and multi-task learning~\cite{li2016deepsaliency,qin2019basnet} have been widely studied.
\textit{Deng et al.} \cite{deng2018r3net} use the low-level and high-level features of FCN to learn residuals between intermediate saliency predictions and ground truth for refining saliency maps.
\textit{Wu et al.} \cite{wu2019cascaded} propose a cascaded partial decoder (CPD) that discards large-resolution features in shallow layers for acceleration, and fuses features in deep layers to obtain accurate saliency maps.
\textit{Liu et al.} \cite{liu2019simple} present pool-based modules to progressively refine features at multiple scales producing detailed results.
The boundary prediction task~\cite{zhou2022edge} captures accurate boundary information of salient objects.
\textit{Qin et al.}~\cite{qin2019basnet} design a hybrid loss for predicting the boundaries of salient objects.
However, boundary supervision lacks the consideration of the interaction between boundary pixels and target pixels.
Inspired by the level set method~\cite{zhang2019resls,cai2021avlsm}, we develop a novel signed distance map auxiliary module (SDMAM) to improve encoder features.
SDMAM takes into account the distance relation of pixels in boundary neighborhoods.
The distance relationship between foreground-background region pixels and boundary pixels can effectively enhance the inter-class separability of features.

\subsection{RGB-T Salient Object Detection}
Compared to RGB images, RGB-T images offer more information of salient objects~\cite{wei2021syncretic}.
In recent years, synergistic SOD between thermal and visible images has been widely studied~\cite{wang2018rgb,tu2019m3s,gao2021unified}.
The dual encoders extract RGB-T features respectively, and the decoder outputs the salient prediction results~\cite{tu2022rgbt}.
The RGB-T SOD methods take full advantage of the complementary capabilities between multimodal sensors to generate cross-modal robust fusion features~\cite{wang2021cgfnet,huo2021efficient,liang2022multi}.
\textit{Tu et al.} \cite{tu2019rgb} suggest a collaborative graph learning algorithm that uses superpixels as graph nodes to learn RGB-T node saliency.
\textit{Zhang et al.} \cite{zhang2019rgb} transform multi-spectral SOD into a CNN feature fusion problem, and propose to capture semantic information and visual details of RGB-T at different depths by fusing multi-level CNN features.
\textit{Tu et al.} \cite{tu2021multi} exploit the complementarity of different modalities of image content and multiple types of cues to extract multi-level multimodal features.
\textit{Zhou et al.} \cite{zhou2021ecffnet} propose an effective and consistent feature fusion network that combines features of different levels through a multi-level consistent fusion module to obtain complementary information.
In this paper, in order to handle long-range dependencies between RGB-T, we develop a cross-spectra fusion transformer.
Furthermore, 
we propose a feature refinement approach with direction field (FRDF) to enhance the intra-class compactness of salient objects.
FRDF exploits feature far from the boundary to refine the features of pixels close to the boundary.

\begin{figure*}[t]  
	\centering 
	\includegraphics[scale=0.55]{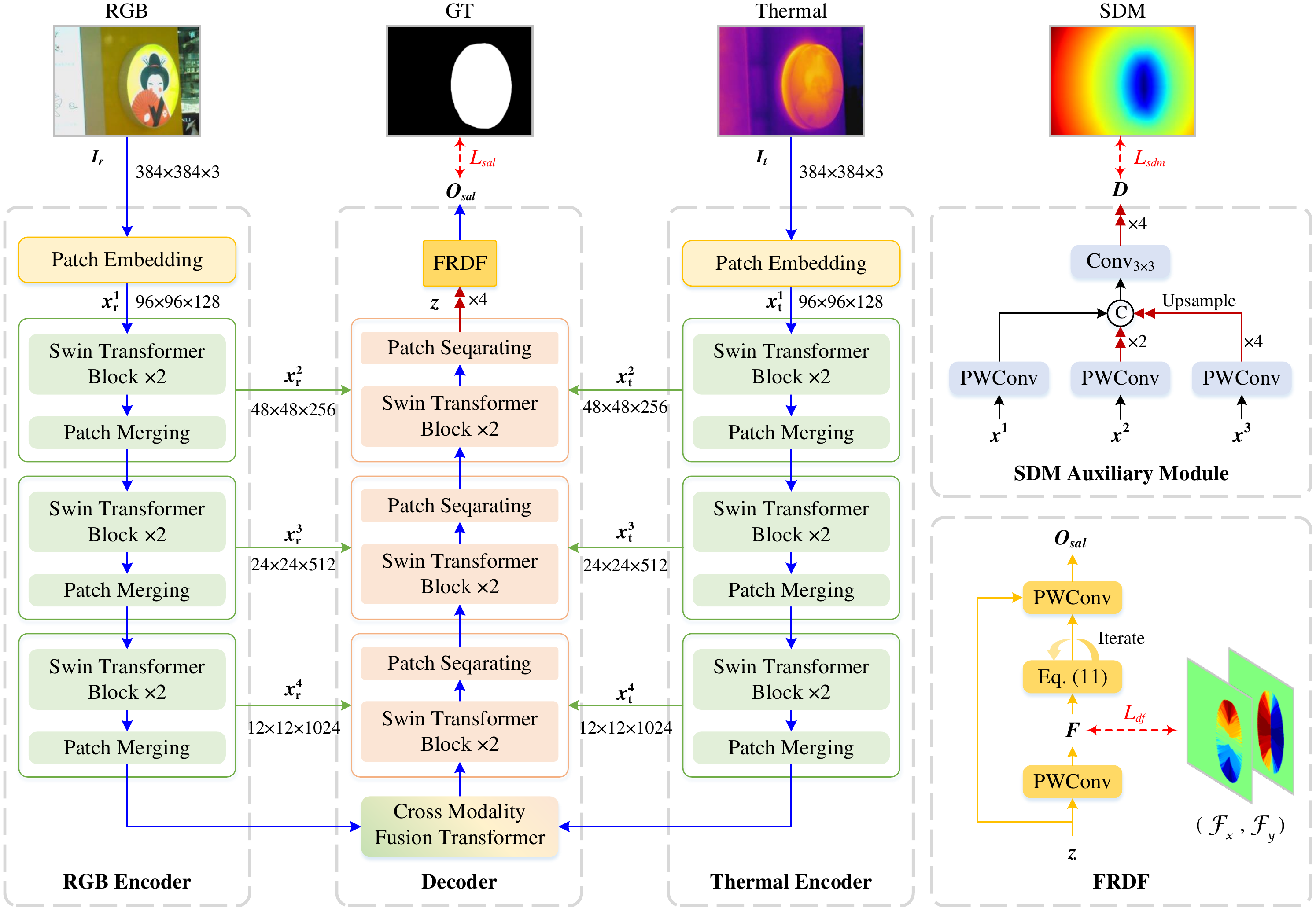}
	\caption{
		The framework of our proposed PRLNet.
		Our network consists of four main parts, 
		namely dual-stream encoders, 
		RGB-T decoder, 
		SDM auxiliary module 
		and a feature refinement approach with directional fields (FRDF).
		First, multiscale features of RGB-T are extracted by a dual-stream swin transformer encoder. (Sec.~\ref{ssec:dual_encoder}).
		Then, we construct SDMAM for encoders to learn the distance relationship between regional pixels and boundary pixels (Sec.~\ref{ssec:sdmam}).
		Next, the reverse swin transformer decoder aggregates the complementarity between different levels of RGB-T features (Sec.~\ref{ssce:decoder}), where $ \copyright $ denotes concatenation.
		In addition, FRDF is further designed by exploring direction information between salient pixels to strengthen the intra-class compactness of the salient features (Sec.~\ref{ssce:frdf}).
		Finally, we propose a novel position-aware relation learning loss to generate object masks with clear boundaries and homogeneous regions (Sec.~\ref{ssec:loss}).
	}
	\label{Fig.model} 
\end{figure*}

\subsection{Swin Transformer}
Compared with CNN, transformer has an advantage in modeling long-range dependencies \cite{zhang2022learning,vaswani2017attention}.
ViT~\cite{dosovitskiy2021an} and DETR~\cite{carion2020end} apply transformer to computer vision tasks and achieve promising performance.
However, the computational complexity of the transformer is proportional to the square of the image size.
To handle high-resolution images, swin transformer~\cite{liu2021swin} introduces the hierarchical structure commonly used in CNN and achieves SOTA results on dense prediction tasks.
Swin transformer gradually becomes a powerful general backbone network for SOD \cite{zeng2022dual}.
\textit{Liu et al.} \cite{Liu2021swinNet} propose a cross-modal fusion network based on the swin transformer for RGB-T SOD, bridging the gap between two modalities through an attention mechanism.
\textit{Zhu et al.} \cite{zhu2022dftr} encode multi-scale features via the swin transformer in a coarse-to-fine manner to learn salient region feature representations.
In this paper, 
swin transformer block is used as the backbone for both the encoder stage and decoder stage.
Specifically, we employ dual-swin transformer encoders to extract multi-scale features from RGB and thermal images, respectively.
Referring to the patch merging layer, we design a patch separating layer to decode RGB-T hierarchical features and generate robust results with multispectral complementarity.

\section{PRLNet}
\label{sec:method}
In this section, we elaborate on our proposed PRLNet for RGB-T SOD with swin transformer.
The overall architecture is illustrated in Fig.~\ref{Fig.model}, which consists of four main parts: 
Dual-stream encoders for both RGB-T images, 
a decoder for pixel-by-pixel prediction, 
a SDM auxiliary module (SDMAM) 
and a feature refinement approach with directional fields (FRDF).
They are simultaneously optimized during the training process.

As shown in Fig.~\ref{Fig.model}, PRLNet takes the RGB-T image pair as input, and segments the precise mask of the salient objects.
We first use the dual-stream swin transformer encoder to generate multi-scale features of RGB and thermal images (Sec.~\ref{ssec:dual_encoder}).
Then, to improve the boundary perception of the encoder, we introduce SDMAM to learn the distance relationship between regional pixels and boundary pixels.
SDMAM enhances the separability of foreground-background features (Sec.~\ref{ssec:sdmam}).
Next, we design a patch separating layer and construct an inverse swin transformer, which aggregates different levels of RGB-T features (Sec.~\ref{ssce:decoder}).
To facilitate the robust cross-spectral features from the decoder,
we further refine them with the direction information between salient pixels to strengthen the intra-class compactness of the salient features (Sec.~\ref{ssce:frdf}).
Finally, benefiting from the effective learning of position relations, we present a position-aware relation learning loss function to generate object masks with clear boundaries and homogeneous regions (Sec.~\ref{ssec:loss}).
The pipeline of PRLNet is illustrated in Algorithm~\ref{alg:algorithm}.

\subsection{Dual-stream Swin Transformer Encoder}
\label{ssec:dual_encoder}
Swin Transformer introduces hierarchical feature mapping and shifted window attention, which has both the advantages of transformer and CNN structure~\cite{liu2021swin}.
We employ two Swin Transformers to extract efficient features for RGB-T image pairs, respectively.
Concretely, the images are first divided into $ 4 \times 4 $ patches and then input to the patch embedding layer, which is a $ 4 \times 4 $ convolution with stride 4.
Next, as shown in Fig.~\ref{Fig.model}, RGB-T salient features are extracted by three swin transformer layers (ST), 
consisting of swin transformer block (STB) and patch merging layer (PM).
That is,
\begin{equation}
	\begin{aligned}
		\mathbf{R} &= \{ \mathbf{x}^{i}_{r} \}^{4}_{i=1}=\mathrm{ST}_{r}\left(\boldsymbol{I}_{r}\right),\\
		\mathbf{T} &= \{ \mathbf{x}^{i}_{t} \}^{4}_{i=1}=\mathrm{ST}_{t}\left(\boldsymbol{I}_{t}\right).
	\end{aligned}
	\label{Eq.encoder}
\end{equation}
The dual encoder outputs the hierarchical representation 
$ \mathbf{R} $ and 
$ \mathbf{T} $, 
where $ \mathbf{x}^{i}_{r} $ and $ \mathbf{x}^{i}_{t} $ denote the $ i $-th layer features of the RGB and thermal encoder, respectively.
$ \boldsymbol{I}_{r} $ and $ \boldsymbol{I}_{t} $ indicate the RGB and thermal images, which are the input of th encoder. The bold symbols indicate the matrix.
The $ \mathrm{ST}_{r} $ and $ \mathrm{ST}_{t} $ functions are composed of STB and PM, and represent the standard swin transformer backbones in RGB and thermal branches, respectively.

\begin{figure}[t]  
	\centering 
	\includegraphics[scale=0.58]{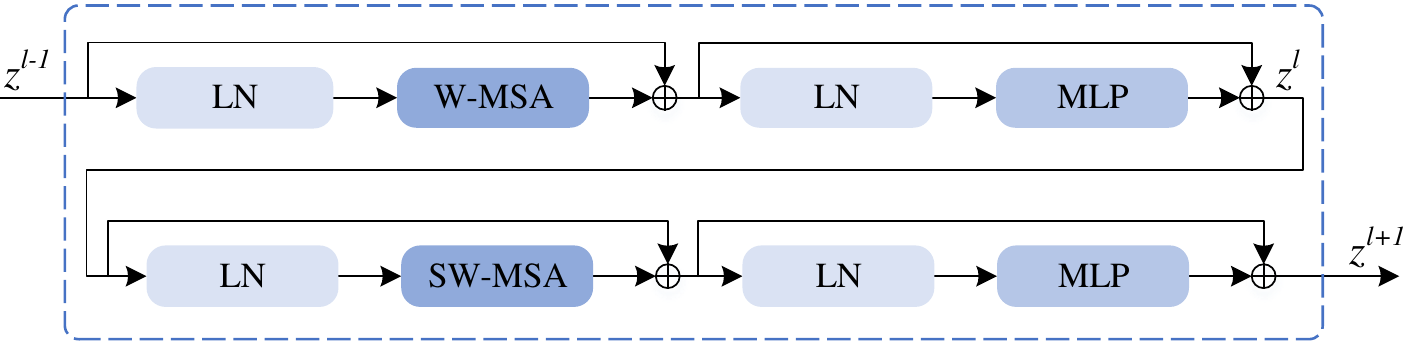}
	\caption{The architecture of the swin transformer block (STB).
		W-MSA calculates the pairwise attention of each token in the window. 
		SW-MSA shifts the window of W-MSA by half the window length.
	}
	\label{Fig.STB} 
\end{figure}

\begin{figure}[t]  
	\centering 
	\subfigure[Patch Merging]
	{\includegraphics[scale=0.25]{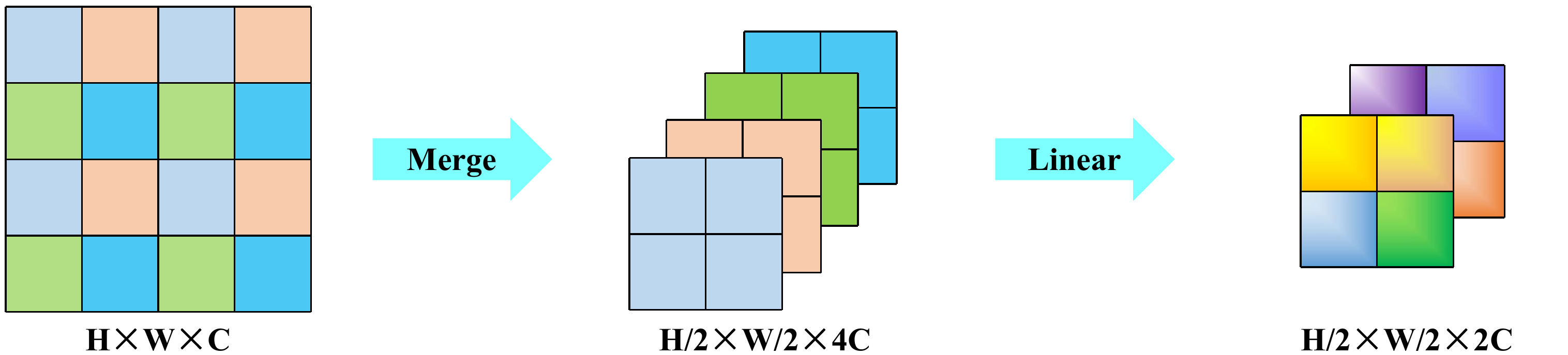}}
	\centering 
	\subfigure[Patch Separating]
	{\includegraphics[scale=0.25]{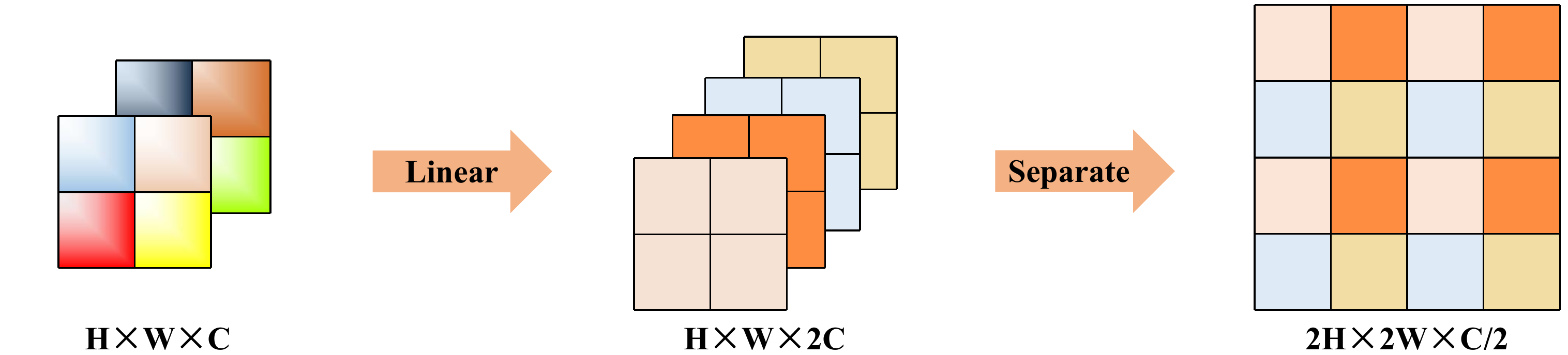}}
	\caption{
		\textbf{(a)} Patch merging layer (PM) merges the neighboring patches into a new patch, thus reducing the resolution.
		\textbf{(b)}~Our proposed patch separating layer (PS) upsamples features by expanding each patch into multiple sub-patches.
	} 
	\label{Fig.PS} 
\end{figure}

Different from ViT block~\cite{dosovitskiy2021an}, STB replaces the multihead attention mechanism (MSA) of ViT with window-based MSA (W-MSA) and shifted window-based MSA (SW-MSA).
More formally, STB is defended as
\begin{equation}
	\begin{aligned}
		\hat{z}^{l}		&=\text{W-MSA}\left(\text{LN}\left(z^{l-1}\right)\right)+z^{l-1}, \\
		z^{l}			&=\text{MLP}\left(\text{LN}\left(\hat{z}^{l}\right)\right)+\hat{z}^{l}, \\
		\hat{z}^{l+1}	&=\text{SW-MSA}\left(\text{LN}\left(z^{l}\right)\right)+z^{l}, \\
		z^{l+1}			&=\text{MLP}\left(\text{LN}\left(\hat{z}^{l+1}\right)\right)+\hat{z}^{l+1}.
	\end{aligned}
\end{equation}
where $ \hat{z}^{l}	 $ and $ z^{l} $ denote the output feature of the (S)W-MSA module and the MLP module for block $ l $, respectively.
Fig.~\ref{Fig.STB} demonstrates the detailed architecture of STB.
STB uses a layernorm (LN) layer before each MSA module and each multilayer perceptron (MLP), followed by residual connections.
As shown in Fig.~\ref{Fig.PS} (a), PM reduces the resolution of the features and increases the number of channels of the features.

\subsection{SDM Auxiliary Module}
\label{ssec:sdmam}

The signed distance map (SDM)~\cite{zhang2019resls,cai2021avlsm} models the distance relationship between pixels in the foreground-background region and the boundary,
and further distinguishes foreground and background with positive and negative signs.
According to the ground truth $ \boldsymbol{G} $, SDM transformation $ \mathcal{D}(p) $ for each pixel $ p \in \boldsymbol{G} $ is given by:
\begin{equation}
	\label{Eq.level_set}
	\mathcal{D}(p)=\left\{
	\begin{array}{ll}
		-\mathop{\inf}\limits_{\forall b \in \partial \mathcal{S}}d(p, b), 	& p \in \mathcal{S}_{\text {sal}} \\  \vspace{1ex}
		0, 																	& p \in \partial \mathcal{S} \\  \vspace{1ex}
		+\mathop{\inf}\limits_{\forall b \in \partial \mathcal{S}}d(p, b), 	& p \in \mathcal{S}_{\text {bg}}
	\end{array}\right.
\end{equation}
where $ \mathrm{inf} $ denotes the infimum, 
$ b $ is the boundary pixel. 
In Eq.~\eqref{Eq.level_set}, $ \partial \mathcal{S} $ is the zero level set and also represents the pixel set of the target boundary.
$ \mathcal{S}_{\text {sal}} $ and $ \mathcal{S}_{\text {bg}} $ indicate the salient object pixel set and background pixel set, respectively.
In our work, $ d(\cdot) $ indicates the Euclidean distance. 
As shown in Fig.~\ref{Fig.supervise_inf}, 
SDM not only perceives the boundary of an object, but also predicts whether the pixel is located inside or outside the object.
For each pixel $ p \in \boldsymbol{G} $, 
the sign of $\mathcal{D}(p)$ indicates whether it is located outside (i.e., $ \mathcal{D}(p)>0 $) or inside (i.e., $ \mathcal{D}(p)<0 $) the object.
$ \mathcal{D}(p)=0 $ denotes the boundary of the object.
$ \left|\mathcal{D}(p)\right| $ represents the distance from pixel $ p $ to the boundary.

In order to precisely perceive the boundaries of salient objects, 
we present a SDM auxiliary module (SDMAM) to learn the distance relation between region pixels and boundary pixels.
Benefiting from SDM, SDMAM can effectively strengthen the inter-class separability of foreground-background region features.
The upper right part of Fig.~\ref{Fig.model} shows the structure of SDMAM in detail.
The shallow high-resolution features contain rich texture information~\cite{liu2019deep}.
SDMAM integrates RGB-T shallow features to predict the distance relationship between pixels.
Formally,
\begin{equation}
	\boldsymbol{D}=\mathrm{SDMAM}\left( \mathbf{x}^{1}, \mathbf{x}^{2}, \mathbf{x}^{3}\right).
	\label{Eq.SDM}
\end{equation}
where $ \mathbf{x}^{i}=\operatorname{concat}\left( \mathbf{x}^{i}_{r}, \mathbf{x}^{i}_{t} \right) $, $ i=1,2,3 $.
$ \boldsymbol{D} \in \mathbb{R}^{h \times w \times 1} $ represents the prediction result of SDMAM.
The dimensions of $ \mathbf{x}^{1} $ , $\mathbf{x}^{2}$ and $\mathbf{x}^{3} $ are $\mathbb{R}^{\frac{h}{4} \times \frac{w}{4} \times c} $, $\mathbb{R}^{\frac{h}{8} \times \frac{w}{8} \times 2c} $ and $\mathbb{R}^{\frac{h}{16} \times \frac{w}{16} \times 4c} $, respectively.
In this paper, $h=w=384$, $c=128$.

Specifically, the multi-scale features $ \{\mathbf{x}^1, \mathbf{x}^2, \mathbf{x}^3 \} $ are further fused by pointwise convolution ($ \operatorname{PWConv} $) \cite{lin2013network} with ReLU
and upsampling operations,
\begin{equation}
	\label{Eq.y}
	\mathbf{y}^{i}=\left[\operatorname{PWConv}\left(\mathbf{x}^{i}\right)\right]^{\times (2^{i-1})},
\end{equation}
where $ i=1,2,3 $. 
$ [\cdot]^{\times (n)} $ denotes upsampling the features by $ n $ times.
In Eq.~\eqref{Eq.y}, the different scale high-resolution features $ \mathbf{y}^{i} \in \mathbb{R}^{\frac{h}{4} \times \frac{w}{4} \times 32} $.
Finally, the multi-scale multi-spectral features $ \mathbf{y} $ are fused by $ 3 \times 3 $ convolution and upsampled to the resolution of the input image.

\begin{equation}
	\begin{aligned}
		\mathbf{y} &=\operatorname{concat}\left(\mathbf{y}^{1}, \mathbf{y}^{2}, \mathbf{y}^{3}\right), \\
		\boldsymbol{D} &=\operatorname{tanh}\left[\operatorname{Conv}_{3 \times 3}(\mathbf{y})\right]^{\times(4)},
	\end{aligned}
\end{equation}
where the output of SDMAM is $ \boldsymbol{D} \in \mathbb{R}^{h \times w \times 1} $,
$ \operatorname{Conv}_{3 \times 3}$ indicate ${3 \times 3} $ convolution with stride $ 1 $.

\subsection{Reverse Swin Transformer Decoder}
\label{ssce:decoder}

Our decoder is designed to decode patches as saliency maps.
Hence, we propose a novel patch upsampling method with multi-level patch fusion and a patch-based SOD decoder.

\subsubsection{Cross Spectrum Fusion Transformer}
Concretely, the RGB-T encoder feature map $ \mathbf{x}^{4} =\operatorname{concat}\left( \mathbf{x}^{4}_{r}, \mathbf{x}^{4}_{t} \right) $ is flattened into an input sequence.
A set of queries $ \mathbf{Q} $, keys $ \mathbf{K} $ and values~$ \mathbf{V} $ is computed by embedding the input sequence into three weight matrices.

In Eq.~\eqref{Eq.z4}, we compute cross-spectral attention $\mathbf{z}^{4}$ as in~\cite{vaswani2017attention}.
\begin{equation}
	\label{Eq.z4}
	\mathbf{z}^{4}=\operatorname{Attention}(\mathbf{Q}, \mathbf{K}, \mathbf{V})=\operatorname{softmax}\left(\frac{\mathbf{Q} \mathbf{K}^{T}}{\sqrt{d}}\right) \mathbf{V},
\end{equation}
where $ \mathbf{z}^{4} \in \mathbb{R}^{\frac{h}{32} \times \frac{w}{32} \times 8c} $,
$ \sqrt{d} $ is an adjustment factor that prevents the $ \operatorname{softmax} $ function from having too large an input value resulting in too small a partial derivative.

\subsubsection{Reverse Swin Transformer Decoder}
In swin transformer, the patch merging (PM) integrates patches of different windows to reduce the spatial resolution of feature maps.
Inspired by PM,
we design patch separating (PS) to upsample patches by separating each patch for multiple sub-patches.
As shown in Fig.~\ref{Fig.PS} (b), 
Based on PS, we propose a reverse swin transformer layer (RST) for the decoder.

The reverse swin transformer decoder is illustrated in the middle of Fig.~\ref{Fig.model}.
RST layer includes STB and PS.
For RGB-T features of encoders,
RST generates more patches and progressively decodes the patches into high-resolution saliency maps, as in Eq.~\eqref{Eq.decoder}.
\begin{equation}
	\label{Eq.decoder}
	\mathbf{z}^{i} =\mathrm{RST}\left(\mathbf{z}^{i+1}, \mathbf{x}^{i+1}\right),
\end{equation}
where $ i=1,2,3 $.
The dimensions of $ \mathbf{z}^{1} $, $\mathbf{z}^{2}$ and $ \mathbf{z}^{3}$ are $\mathbb{R}^{\frac{h}{4} \times \frac{w}{4} \times c} $, $\mathbb{R}^{\frac{h}{8} \times \frac{w}{8} \times 2c} $ and $\mathbb{R}^{\frac{h}{16} \times \frac{w}{16} \times 4c} $, respectively.
The salient swin transformer decoder output $ \mathbf{z} \in \mathbb{R}^{h \times w \times 64}$ is obtained by upsampling $ \mathbf{z}^{1} $ by a factor of 4.

\subsection{Feature Refinement Approach with Direction Field}
\label{ssce:frdf}

The direction field (DF)\cite{zhang2022discriminative} offers the directional relationship between salient pixels.
The direction vector of each pixel points from the boundary to the center.
The mathematical definition of the direction field function $ \mathcal{F} $ is shown in Eq.~\eqref{Eq.direction_field}.
The direction of $ \mathcal{F}(p) $ is from $ b $ pointing to $ p $, and $ b $ is the nearest pixel to $ p $ on the boundary.
For the pixel $ p \in \boldsymbol{G} $,
\begin{equation}
	\label{Eq.direction_field}
	\mathcal{F}(p)=\left\{
	\begin{array}{cl}
		\mathop{\inf}\limits_{\forall b \in \partial \mathcal{S}}{\vec{bp}},  & p \in \mathcal{S}_{\text {sal}} \vspace{1ex} \\ 
		(0,0). 					& p \in \mathcal{S}_{\text {bg}} 
	\end{array}\right.
\end{equation}
where $ \mathcal{S}_{\text {sal}} $ and $ \mathcal{S}_{\text {bg}} $ denote the salient object pixel set and background pixel set, respectively.

The refinement of the initial predicted features provides an effective way to improve salient object masks.
Based on this idea, we design a feature refinement approach with directional field (FRDF). FRDF uses features inside the object to improve the feature representation near the boundary with the help of directional information between pixels.
FRDF progressively enforces the intra-class compactness of salient region features through several iterative updates.
As shown in the bottom right of Fig.~\ref{Fig.model},
we first use the decoder feature $\mathbf{z}$ to predict the directional field feature $ \boldsymbol{F} \in \mathbb{R}^{h \times w \times 2}$ in Eq.~\eqref{Eq.DF}.
\begin{equation}
	\label{Eq.DF}
	\boldsymbol{F} =\operatorname{PWConv}\left(\mathbf{z}\right).
\end{equation}
Then, the initial predicted saliency feature map is refined step by step iteratively according to Eq.~\eqref{Eq.zk}.
\begin{equation}
	\label{Eq.zk}
	\mathbf{z}_{k}(p)=\mathbf{z}_{k}\left(p_{x}+\boldsymbol{F}_{x}(p), p_{y}+\boldsymbol{F}_{y}(p)\right),
\end{equation}
where
$ \mathbf{z}_{k} $ denotes the salient feature map after the $ k $-th iteration.
The number of iterations is setting as $ K = 5 $, which is further ablated with experiments in Sec.~\ref{sssec:hyperpara}. 
$ p_x $ and $ p_y $ indicate the $ x $ and $ y $ coordinates of pixel $ p $, respectively.
The output of FRDF is the refined feature $ \mathbf{z}^{*} \in \mathbb{R}^{{h} \times {w} \times 2c}$.

Finally, the $ \operatorname{PWConv} $ layer combines initial feature $ \mathbf{z} $ with the rectified feature $ \mathbf{z}^{*}  $ to generate the salient mask $ \boldsymbol{O}_{sal} \in \mathbb{R}^{{h} \times {w} \times 1} $.
Both SDM and DF prediction are supervised, 
which will be discussed in Sec.~\ref{ssec:loss}.
Based on the ground truth $ \boldsymbol{G} $, we obtain the true supervised signal for the SDMAM and FRDF.

\subsection{Loss Function}
\label{ssec:loss}
According to the ground truth $ \boldsymbol{G} $ of the image,
the true SDM and the true direction field of the salient object can be calculated by mathematical models, $ i.e. $, Eq. \eqref{Eq.level_set} and Eq. \eqref{Eq.direction_field}.
\begin{equation}
	\begin{aligned}
		{\boldsymbol{D}_{gt}} &= \mathcal{D}(\boldsymbol{G}) \\
		{\boldsymbol{F}_{gt}} &= \mathcal{F}(\boldsymbol{G})
	\end{aligned}
	\label{Eq.D_F}
\end{equation}
In Eq. \eqref{Eq.D_F}, $ {\boldsymbol{D}_{gt}} $ is the true SDM and $ {\boldsymbol{F}_{gt}} $ is the true DF.
They guide SDMAM and FRDF to enhance intra-class compactness and inter-class separability, which are weakly supervision for RGB-T SOD.

\subsubsection{SDM Loss}
\label{sssec:sdm_loss}
SDM loss is
\begin{equation}
	\label{Eq.Lsdm}
	\mathcal{L}_{sdm}=\sum_{p \in \Omega}\left\|\boldsymbol{D}-{\boldsymbol{D}_{gt}} \right\|^{2},
\end{equation}
where $ \Omega $ denotes all pixels, $ \boldsymbol{D} $ is the predicted result of SDMAM.
$ \mathcal{L}_{sdm} $ drives PRLNet to learn the distance relationship between foreground-background regions and boundaries, effectively enhancing the inter-class differences of salient features.

\subsubsection{Direction Field Loss}
\label{sssec:df_loss}
DF loss is
\begin{equation}
	\label{Eq.loss_df}
	\mathcal{L}_{df}=\sum_{p\in \Omega} \left(\|\boldsymbol{F}-\hat{\boldsymbol{F}}\|_{2} + \left\|\cos ^{-1}\langle \boldsymbol{F}, {\boldsymbol{F}_{gt}}\rangle\right\|^{2}\right),
\end{equation}
where $ \boldsymbol{F} $ and $ {\boldsymbol{F}_{gt}} $ indicate the predicted DF and the corresponding ground truth, respectively.
$ \mathcal{L}_{df} $ guides the model to learn the directional relationship between pixels, which rectifies features of boundary neighborhood by exploiting the features inside salient objects.

\subsubsection{Direction-aware Smoothness Loss}
We develop a novel direction-aware smoothness loss ($ \mathcal{L}_{DS} $) that enhances the compactness of regions and the boundary clearness. 
We calculate the first order derivative of the saliency map in the smooth term~\cite{godard2017unsupervised,tu2021multi}.
$ \mathcal{L}_{DS} $ is defined as follows,
\begin{equation}
	\label{Eq.LDS}
	\mathcal{L}_{DS}(\boldsymbol{O},\boldsymbol{G})=\sum_{p \in \Omega} \sum_{\partial_{x,y}} w\left(p\right) \psi\left(\left|\partial \boldsymbol{O}\right| e^{-\alpha \left|\partial \boldsymbol{G}\right|}\right),
\end{equation}
\begin{equation}
	\label{Eq.w}
	w(p)=\left\{
	\begin{array}{cl}
		{\| \mathcal{F}(p) \|}^{-1}, & p \in \mathcal{S} \vspace{1ex} \\ 
		1, 					& p \in \mathcal{S}_{\text {bg}}
	\end{array}\right.
\end{equation}
where $\psi(m)=\sqrt{m^{2}+{0.001}^2}$, $ \boldsymbol{O} $ and $ \boldsymbol{G} $ represent the predicted salient result and ground truth, respectively.
$ \partial_{x,y} $ denotes the partial derivatives in $ x $ and $ y $ directions.
Same as~\cite{wang2018occlusion}, we set $ \alpha =10$ to balance the contribution of the edges.
In Eq.~\eqref{Eq.w}, $ w\left(p\right) $ indicates the weight on pixel $ p $.
Therefore, the saliency loss is
\begin{equation}
	\label{Eq.loss_sal}
	\mathcal{L}_{sal}=\mathcal{L}_{DS}\left(\boldsymbol{O}_{sal}, \boldsymbol{G}\right).
\end{equation}

Finally, our position-aware relation learning (PRL) loss is
\begin{equation}
	\label{Eq.loss_all}
	\mathcal{L}_{prl} = \mathcal{L}_{sal} + \lambda_{1} \mathcal{L}_{sdm} + \lambda_{2} \mathcal{L}_{df},
\end{equation}
where $ \lambda_{1} $ and $ \lambda_{2} $ are the hyper-parameters balancing the contributions of the two losses,
which are set via ablative analysis in Sec.~\ref{sssec:hyperpara}.
The proposed PRLNet is optimized through Eq.~\eqref{Eq.loss_all} jointly.
Our proposed position-aware relation learning loss can effectively guide the network to pay more attention to the pixels around the object boundary, 
thereby helping the network to predict salient masks with sharp boundaries and homogeneous regions.

\begin{algorithm}[t]
	\caption{The pipeline of PRLNet}
	\label{alg:algorithm}
	\textbf{Input}: RGB-T images $\{ \boldsymbol{I}_{r}, \boldsymbol{I}_{t} \} $, ground truth \textit{\textbf{G}} \\
	\textbf{Output}: Salient object mask $ \boldsymbol{O}_{sal} $, signed distance map $ \boldsymbol{D} $, direction field $ \boldsymbol{F} $\\
	\begin{algorithmic}[1] 
		\STATE Init true SDM $ {\boldsymbol{D}_{gt}} \leftarrow \mathcal{D}(\boldsymbol{G}) $ using Eq.~\eqref{Eq.level_set}
		\STATE Init true direction field $ {\boldsymbol{F}_{gt}} \leftarrow \mathcal{F}(\boldsymbol{G}) $ using Eq.~\eqref{Eq.direction_field}
		\STATE Init the number of FRDF iterations $ K=5 $
		\WHILE{$ epoch < N$}
		\STATE Extract RGB-T features $ \mathbf{R} $ and $ \mathbf{T} $ using Eq.~\eqref{Eq.encoder}
		\STATE Generate signed distance map $ \boldsymbol{D} $ using Eq.~\eqref{Eq.SDM}
		\STATE Generate decoder output feature $ \mathbf{z} $ using Eq.~\eqref{Eq.decoder}
		\STATE Generate direction field $ \boldsymbol{F} $ using Eq.~\eqref{Eq.DF}
		\FOR{$ k \le K $}
		\STATE Iterative refinement of decoder feature \\ $ \mathbf{z}_{} \leftarrow \mathbf{z}_{} + \boldsymbol{F} $ using Eq.~\eqref{Eq.zk}
		\ENDFOR
		\STATE Generate salient mask result $ \boldsymbol{O}_{sal} $ using initial features~$ \mathbf{z} $ and refined features $ \mathbf{z}^{*} $
		\STATE Calculate the SDM loss \\ $ \mathcal{L}_{sdm} \leftarrow \mathcal{L}(\boldsymbol{D},{\boldsymbol{D}_{gt}})$ using Eq.~\eqref{Eq.Lsdm}
		\STATE Calculate the DF loss \\$ \mathcal{L}_{df} \leftarrow \mathcal{L}(\boldsymbol{F},{\boldsymbol{F}_{gt}})$ using Eq.~\eqref{Eq.loss_df}
		\STATE Calculate the direction-aware smoothness loss $ \mathcal{L}_{sal} \leftarrow \mathcal{L}(\boldsymbol{O}_{sal},{\boldsymbol{G}})$ using Eq.~\eqref{Eq.loss_sal}
		\STATE Calculate the overall loss function of PRLNet $ \mathcal{L}_{prl} \leftarrow \mathcal{L}_{sal} + \lambda_{1} \mathcal{L}_{sdm} + \lambda_{2} \mathcal{L}_{df}, $
		\STATE $ \boldsymbol{O}_{sal}, \boldsymbol{D}, \boldsymbol{F} \leftarrow\arg\min \mathcal{L}_{prl}$
		\ENDWHILE
		\STATE \textbf{return} $ \boldsymbol{O}_{sal} $
	\end{algorithmic}
\end{algorithm}

\section{Experiments}
\label{sec:exp}

In this section, we first introduce the three RGB-T datasets, implementation details, and evaluation metrics.
We then give the details of our experiments.
In particular,
we evaluate our method on three widely used datasets to compare with SOTA methods.
Moreover, ablation studies are also conducted to further validate the validity of our network.

\subsection{Experimental Setup}
\subsubsection{Datasets}
There are three available benchmark datasets for RGB-T SOD tasks,
including 
VT821~\cite{wang2018rgb},
VT1000~\cite{tu2019rgb} and 
VT5000~\cite{tu2022rgbt},
which have 821, 1000, and 5000 aligned image pairs, respectively.
Compared with VT821, VT1000 dataset has more images and scenes, and the quality of the thermal images is better.
VT5000 provides a large-scale dataset for TGB-T SOD.
In addition, VT5000 does not require manual RGB-T image pair alignment, which reduces the errors caused by manual alignment.
VT5000 contains a variety of complex scenes with diverse objects and covers 13 challenges of RGB-T SOD~\cite{tu2022rgbt}, the details are shown in TABLE~\ref{tab:challeng}.
As reported in TABLE~\ref{tab:challeng}, VT5000 simulates image saliency detection under real-world conditions mainly in terms of 
target diversity (BSO, SSO, MSO, CB, CIB, OF, SA and TC), scene complexity (IC, LI and BW) and spectral effectiveness (RGB and T).

\begin{table}[t]
	\centering
	\caption{Details of 13 challenges in VT5000 dataset.}
	\label{tab:challeng}
	\setlength{\tabcolsep}{0.5mm}{%
		\begin{tabular}{cp{7cm}}
			\toprule
			\textbf{Challenge} & \textbf{Describe}                                                                                         \\ \midrule
			BSO                & Big salient object: the proportion of pixels of salient objects to the image is more than 0.26.           \\
			SSO                & Small salient object: the percentage of the number of salient pixels is less than 0.05.                   \\
			MSO                & Multiple salient objects.                                                                                 \\
			CB                 & Center bias: the salient object is out of the center of the image.                                        \\
			CIB                & Cross image boundary: a part of the salient object is outside the image.                                  \\
			OF                 & Out of focus: out of focus causes the whole image to be blurred.                                          \\
			SA                 & Similar appearance: the salient object is similar to the color and texture of the background.			   \\
			TC                 & Thermal crossover: the salient object is similar to its surrounding temperature.                          \\
			IC                 & Image clutter: the scene is cluttered.                                                                    \\
			LI                 & Low illumination: the scene is cloudy or at night.                                                        \\
			BW                 & Bad weather: the scene is rainy or foggy.                                                                 \\
			RGB                & Objects are not clear in RGB image.                                                                       \\
			T                  & Objects are not clear in the thermal image.                                                               \\ \bottomrule
		\end{tabular}%
	}
\end{table}

\subsubsection{Implementation Details}
To extract multispectral features, we initialize our backbone networks through the parameters of the pre-trained Swin-B model \cite{liu2021swin}. 
The whole network is then trained on a large-scale dataset with the proposed position-aware relation learning loss in an end-to-end manner.

For a fair comparison, we use the same setting as in~\cite{deng2018r3net,tu2019rgb,Liu2021swinNet,tu2022rgbt} where half of VT5000 dataset is applied as the training set. 
VT821, VT1000 and the other half of VT5000 are treated as the test set.
In addition, each image is random flipping, cropping and rotation ($ -15^\circ \sim 15^\circ$), and then resized to $ 384 \times 384 $.
We train our models by using adaptive optimizer Adam.
The initial learning rate of the network is set to $10 ^ {-5} $ and is decayed by 0.1 every 100 epochs.
The total epoch number is set to 300.
The mini-batch size is set as 6. 
Our framework is implemented by PyTorch.
The experiment is conducted on a computer with 3.0 GHz CPU, 128 GB RAM, and four NVIDIA GeForce RTX 3090 GPUs.

\subsection{Evaluation Metrics}
In order to facilitate the comparison of the performance of different RGB-T methods,
we use the evaluation metrics commonly used in SOD model:
P-R curves~\cite{zhang2020dense},
S-measure ($ \mathrm{S}_{\alpha} \uparrow $)~\cite{fan2017structure},
F-measure ($ \mathrm{F}_{\beta} \uparrow $)~\cite{achanta2009frequency},
E-measure ($ \mathrm{E}_{m} \uparrow $)~\cite{fan2018enhanced} and 
MAE ($ \mathcal{M} \downarrow$)~\cite{perazzi2012saliency}.
$ \uparrow $ and $ \downarrow $ indicate that the higher the better
and the lower the better, respectively.
The P-R curves and $ \mathrm{F}_{\beta} $ evaluate the quality of the prediction results in terms of Precision and Recall.
$ \mathrm{S}_{\alpha} $ and $ \mathrm{E}_{m} $ mainly measure the structural similarity between predicted saliency mask and GT.
$ \mathcal{M} $ counts the error of the prediction result pixel by pixel.
We use the above metrics to evaluate the model accurately and comprehensively.
The formal definition is as follows.

\textbf{P-R curves.}
We first demonstrate the performance of our model through standard P-R curves~\cite{zhang2020dense}.
Different thresholds ($ [0,255] $) are applied to the prediction to generate binarized result that produces pairs of $ \mathrm{Precision} $-$ \mathrm{Recall} $ values. 
A set of thresholds provides the P-R curve of the model.
Formally, the $ \mathrm{P} $ and $ \mathrm{R} $ are defined based on the binarized salient object mask and the corresponding ground truth in Eq.~\eqref{Eq.PR}.
\begin{equation}
	\label{Eq.PR}
	\mathrm{P}=\frac{\mathrm{TP}}{\mathrm{TP}+\mathrm{FP}}, \quad \mathrm{R}=\frac{\mathrm{TP}}{\mathrm{TP}+\mathrm{FN}},
\end{equation}
where $ \mathrm{TP} $, $ \mathrm{FP} $ and $ \mathrm{FN} $ denote true positive, false positive and false negative, respectively.

\textbf{S-measure.}
The structure measure ($ \mathrm{S}_{\alpha} $) can effectively evaluate the spatial structure compactness between prediction and ground truth~\cite{fan2017structure}.
\begin{equation}
	\label{Eq.Sm}
	\mathrm{S}_{\alpha}=\alpha \mathrm{S}_{o}+(1-\alpha)\mathrm{S}_{r},
\end{equation}
where $ \alpha $ is set as $ 0.5 $ empirically~\cite{fan2017structure}.
In Eq.~\eqref{Eq.Sm}, $ \mathrm{S}_{\alpha} $ integrates object-aware structural similarity $ \mathrm{S}_o $ and region-aware structural similarity $ \mathrm{S}_r $.

\textbf{F-measure.} 
$ \mathrm{F}_{\beta} $ takes into account precision and recall~\cite{achanta2009frequency}, 
and calculates the weighted harmonic mean of $ \mathrm{P} $ and $ \mathrm{R} $:
\begin{equation}
	\mathrm{F}_{\beta}=\frac{\left(1+\beta^{2}\right)\times\mathrm{P} \times \mathrm{R}}{\beta^{2}\times\mathrm{P}+\mathrm{R}},
\end{equation}
where we set $ \beta^{2}=0.3 $ to weigh precision more than recall.

\textbf{E-measure.}
The enhanced-alignment measure metric ($ \mathrm{E}_{m} $) considers both local pixel values and image-level averages.
$ \mathrm{E}_{m} $ captures image-level statistics and local pixel matching information~\cite{fan2018enhanced}.
\begin{equation}
	\label{Eq.Em}
	\mathrm{E}_{m}=\frac{1}{HW} \sum_{i=1}^{H} \sum_{j=1}^{W} \phi_{ij}.
\end{equation}
In Eq.~\eqref{Eq.Em}, $ H $ and $ W $ are the height and width of the object map, respectively.
$ \phi $ is the enhanced alignment matrix~\cite{fan2018enhanced}.

\textbf{MAE.}
The mean absolute error ($ \mathcal{M} $)~\cite{perazzi2012saliency} measures the difference between 
saliency prediction $\boldsymbol{O} \in[0,1]^{H \times W}$ and ground truth mask $\boldsymbol{G} \in\{0,1\}^{H \times W}$,

\begin{equation}
	\mathcal{M}=\frac{1}{HW} \sum_{i=1}^{H} \sum_{j=1}^{W}|\boldsymbol{O}_{ij}-\boldsymbol{G}_{ij}|.
\end{equation}

\begin{table*}[t]
	\centering
	\caption{
		Quantitative comparison with SOTA method on three benchmark datasets in terms of 
		S-measure ($ \mathrm{S}_{\alpha} \uparrow $),
		F-measure ($ \mathrm{F}_{\beta} \uparrow $),
		E-measure ($ \mathrm{E}_{m} \uparrow $) and 
		MAE ($ \mathcal{M} \downarrow$).
		$ \uparrow $ and $ \downarrow $ represent the higher the better and the lower the better, respectively.
		The best result in each column is in {\color{red} \textbf{red}}, and the second is in {\color{blue} blue}.}
	\label{tab:sota}
	\setlength{\tabcolsep}{3.5mm}{%
		\begin{tabular}{@{}ccccccccccccc@{}}
			\toprule
			& \multicolumn{4}{c}{VT821}                                                                                                                                                   & \multicolumn{4}{c}{VT1000}                                                                                                                                                  & \multicolumn{4}{c}{VT5000}                                                                                                                                                  \\ \cmidrule(lr){2-5} \cmidrule(lr){6-9}  \cmidrule(lr){10-13}
			\multirow{-2}{*}{Methods} & \multicolumn{1}{l}{$S_{\alpha} \uparrow$} & \multicolumn{1}{l}{$F_{\beta} \uparrow$} & \multicolumn{1}{l}{$E_{m} \uparrow$}  & \multicolumn{1}{l}{$\mathcal{M} \downarrow$} & \multicolumn{1}{l}{$S_{\alpha} \uparrow$} & \multicolumn{1}{l}{$F_{\beta} \uparrow$} & \multicolumn{1}{l}{$E_{m} \uparrow$}  & \multicolumn{1}{l}{$\mathcal{M} \downarrow$} & \multicolumn{1}{l}{$S_{\alpha} \uparrow$} & \multicolumn{1}{l}{$F_{\beta} \uparrow$} & \multicolumn{1}{l}{$E_{m} \uparrow$}  & \multicolumn{1}{l}{$\mathcal{M} \downarrow$} \\ \midrule
			R3Net                     & 0.785                                     & 0.809                                    & 0.660                                 & 0.073                                        & 0.842                                     & 0.859                                    & 0.761                                 & 0.055                                        & 0.757                                     & 0.790                                    & 0.615                                 & 0.083                                        \\
			CPD                      & 0.818	& 0.718	& 0.843	& 0.079	& 0.907	& 0.863	& 0.923	& 0.031	& 0.855	& 0.787	& 0.894	& 0.046 \\
			PoolNet                   & 0.751                                     & 0.739                                    & 0.578                                 & 0.109                                        & 0.834                                     & 0.813                                    & 0.714                                 & 0.067                                        & 0.769                                     & 0.755                                    & 0.588                                 & 0.089                                        \\
			SGDL                      & 0.765                                     & {\color[HTML]{3531FF} 0.847}                                    & 0.731                                 & 0.085                                        & 0.787                                     & 0.856                                    & 0.764                                 & 0.090                                        & 0.750                                     & 0.824                                    & 0.672                                 & 0.089                                        \\
			ADF   & 0.810	& 0.716	& 0.842	& 0.077	& 0.910	& 0.847	& 0.921	& 0.034	& 0.863	& 0.778	& 0.891	& 0.048 \\
			FMCF                      & 0.760                                     & 0.796                                    & 0.640                                 & 0.080                                        & 0.873                                     & {\color[HTML]{3531FF} 0.899}                                    & 0.823                                 & 0.037                                        & 0.814                                     & 0.864                                    & 0.734                                 & 0.055                                        \\
			MIDD					  & 0.871	& 0.804	& 0.895	& 0.045	& 0.915	& 0.882	& 0.933	& 0.027	& 0.867	& 0.801	& 0.897	& 0.043 \\
			ECFFNet                   & 0.877                                     & 0.810                                    & 0.902                                 & 0.034                                        & 0.923                                     & 0.876                                    & 0.930                                 & 0.021                                        & 0.874                                     & 0.806                                    & 0.906                                 & 0.038                                        \\
			SwinNet                   & {\color[HTML]{3531FF} 0.904}                                 & {\color[HTML]{3531FF} 0.847}                                    & {\color[HTML]{3531FF} 0.926}          & {\color[HTML]{3531FF} 0.030}                                        & {\color[HTML]{3531FF} 0.938}                                     & 0.896                                    & {\color[HTML]{3531FF} 0.947}                                 & {\color[HTML]{3531FF} 0.018}                                        & {\color[HTML]{3531FF} 0.912}                                     & {\color[HTML]{3531FF} 0.865}                                    & {\color[HTML]{3531FF} 0.942}                                 & {\color[HTML]{3531FF} 0.026}                                        \\ \midrule
			Our         & {\color[HTML]{FE0000} \textbf{0.917}}     & {\color[HTML]{FE0000} \textbf{0.860}}             & {\color[HTML]{FE0000} \textbf{0.932}} & {\color[HTML]{FE0000} \textbf{0.025}}        & {\color[HTML]{FE0000} \textbf{0.944}}     & {\color[HTML]{FE0000} \textbf{0.902}}                                    & {\color[HTML]{FE0000} \textbf{0.951}} & {\color[HTML]{FE0000} \textbf{0.016}}        & {\color[HTML]{FE0000} \textbf{0.921}}     & {\color[HTML]{FE0000} \textbf{0.875}}             & {\color[HTML]{FE0000} \textbf{0.948}} & {\color[HTML]{FE0000} \textbf{0.023}}        \\ \bottomrule
		\end{tabular}%
	}
\end{table*}

\begin{figure*}[t]  
	\centering 
	{\includegraphics[scale=0.24]{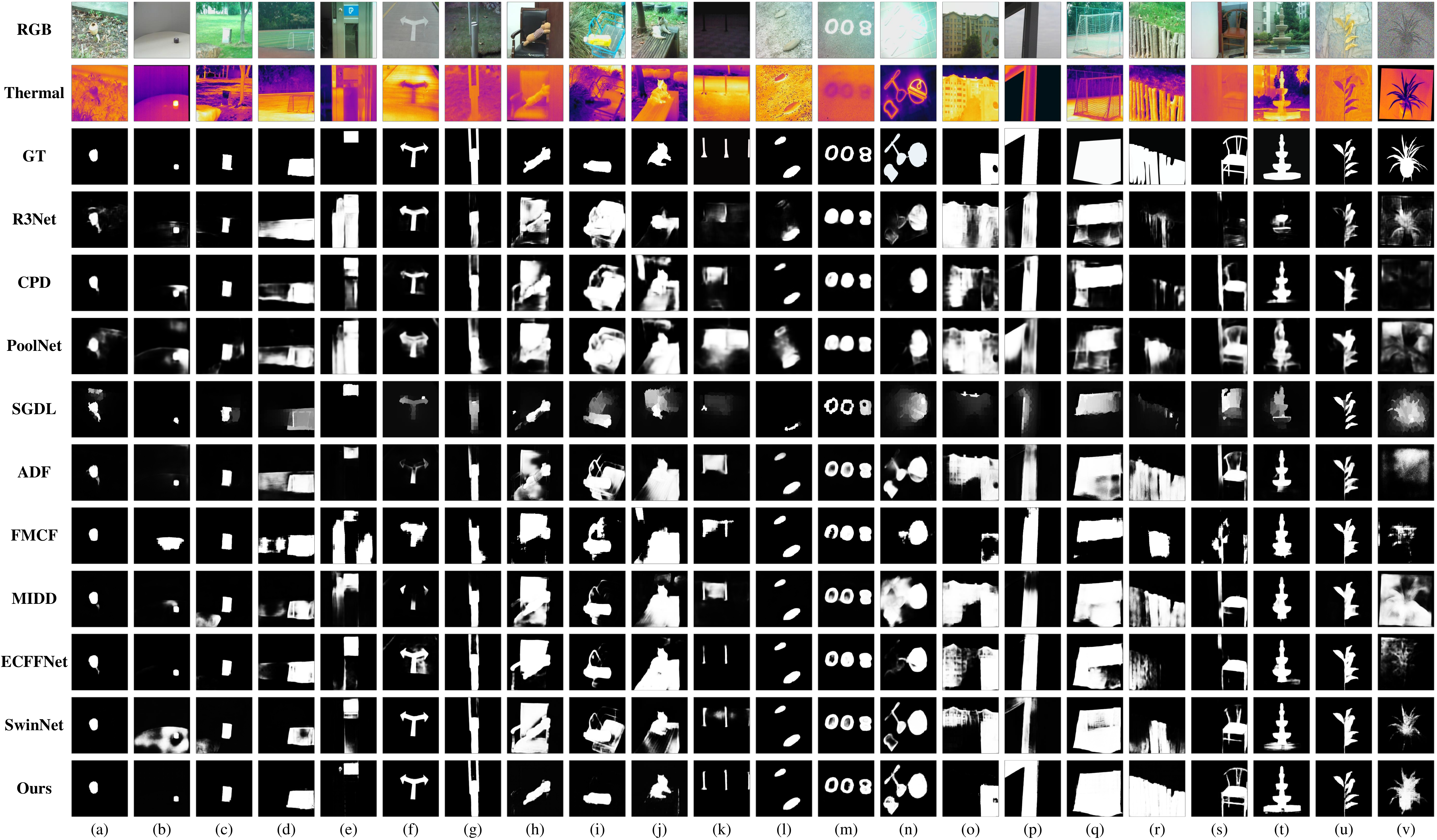}}
	\caption{
		Visual comparisons of different SOTA methods under various challenges, where each column indicates one input image.
		This figure shows that our proposed method (Ours) consistently generates saliency maps close to the Ground Truth (GT).
	}
	\label{Fig.res} 
\end{figure*}

\subsection{Comparison with State-of-the-Art Methods}
\begin{figure*}[t]  
	\centering 
	\subfigure[VT821 dataset]
	{\includegraphics[scale=0.35]{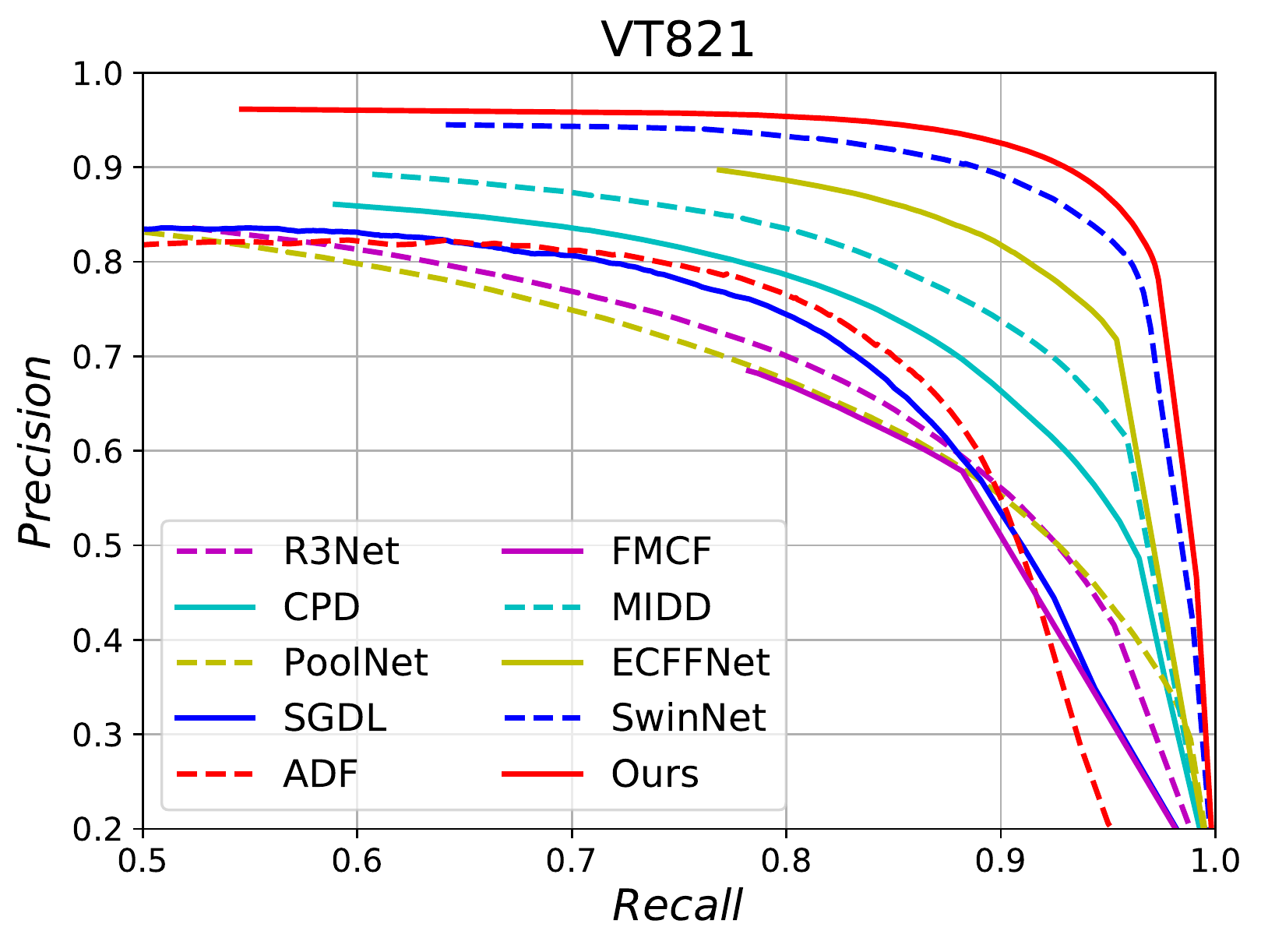}}
	\centering 
	\subfigure[VT1000 dataset]
	{\includegraphics[scale=0.35]{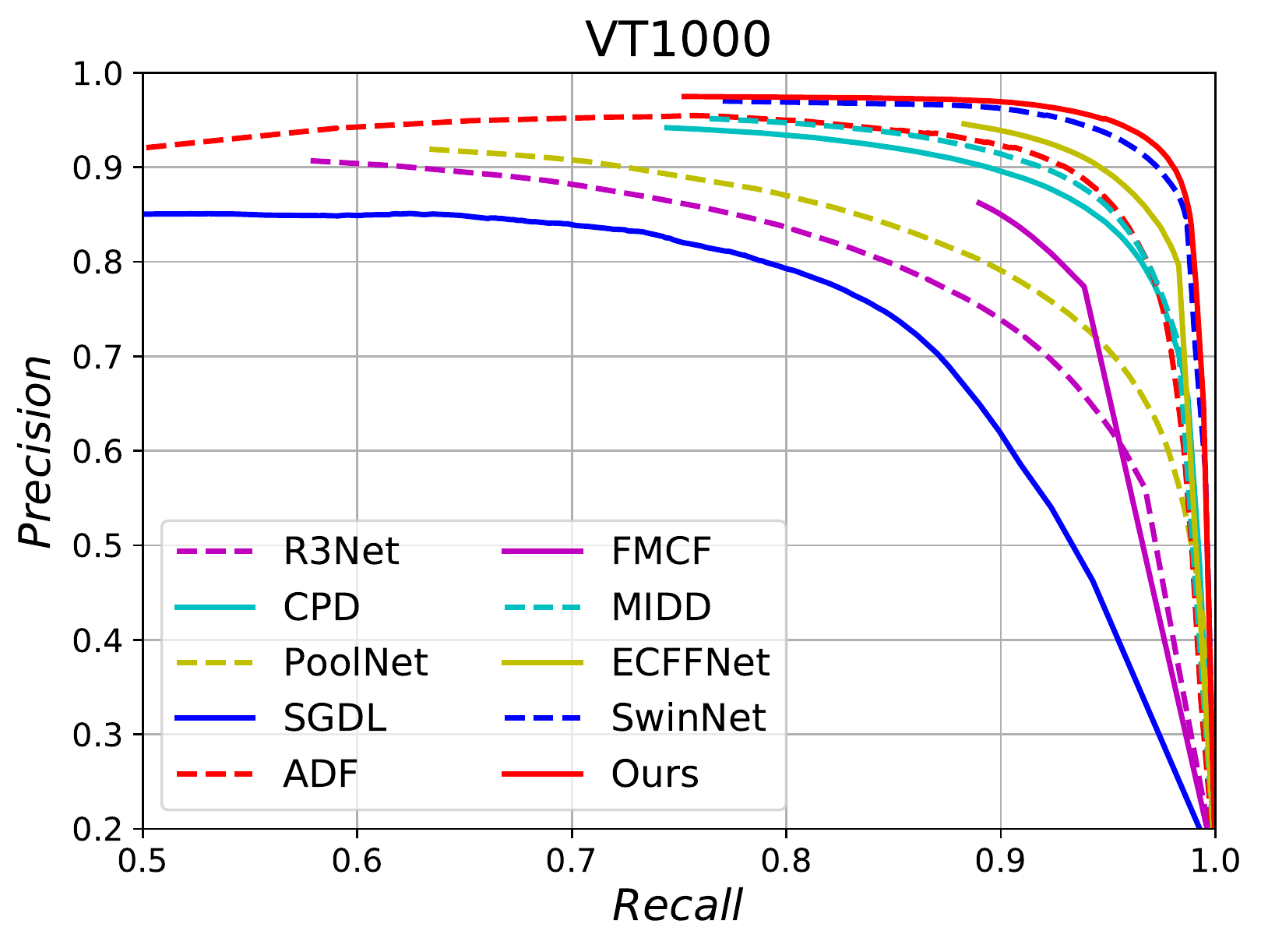}}
	\centering 
	\subfigure[VT5000 dataset]
	{\includegraphics[scale=0.35]{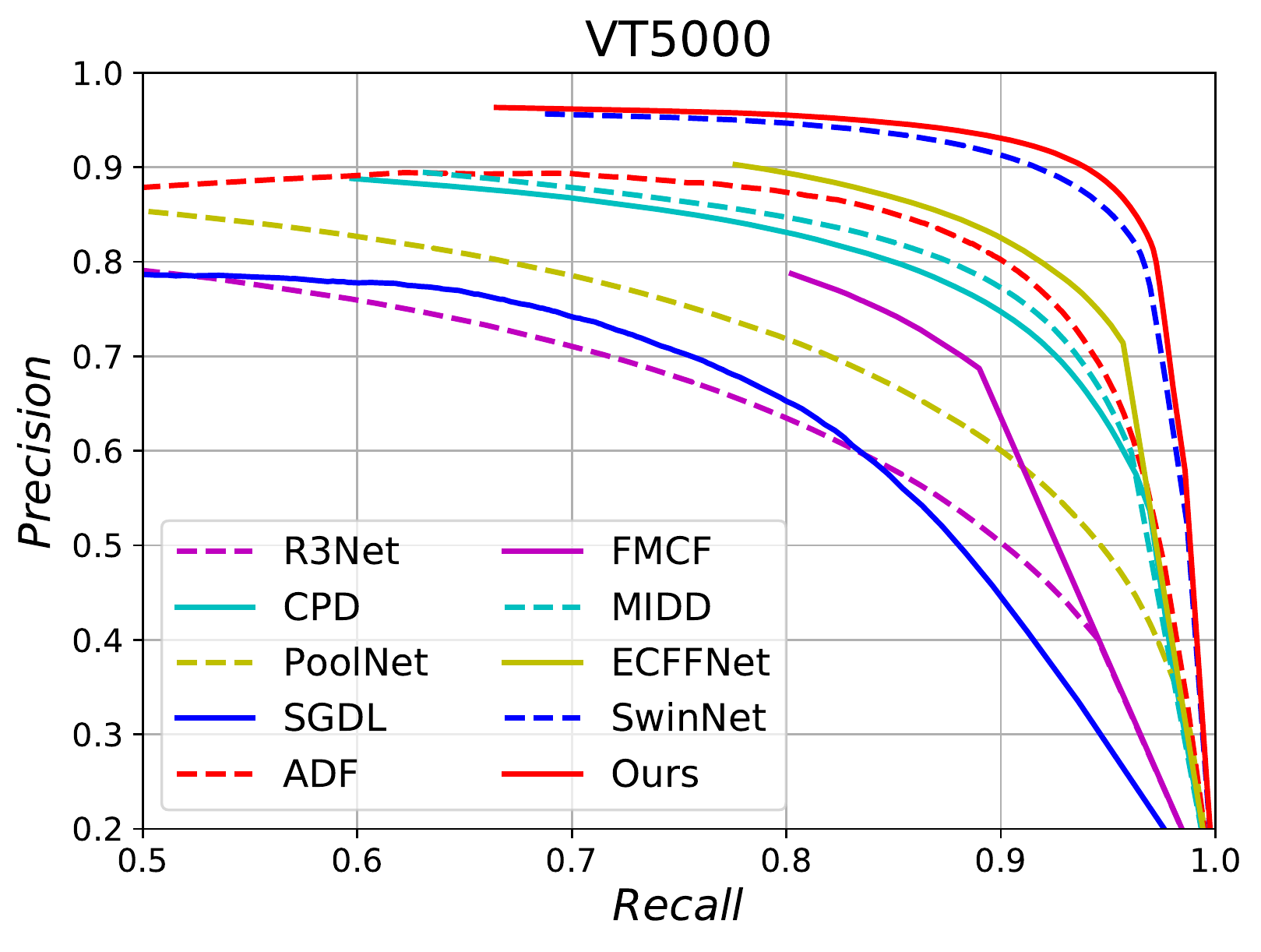}}
	\caption{
		P-R curves comparison of different methods on VT821, VT1000 and VT5000 datasets.
		The P-R curves show that our PRLNet (Ours) consistently outperforms the SOTA models on three datasets.
	} 
	\label{Fig.PR} 
\end{figure*}

To evaluate the validity of the proposed PRLNet, we conduct experiments compared with state-of-the-art methods on three datasets, 
which are shown in TABLE~\ref{tab:sota} and Fig.~\ref{Fig.PR}.
Three RGB SOD methods include 
R3Net~\cite{deng2018r3net}, 
CPD~\cite{wu2019cascaded} and 
PoolNet~\cite{liu2019simple}. 
Six RGB-T SOD methods include
SGDL~\cite{tu2019rgb}, 
ADF~\cite{tu2022rgbt}, 
FMCF~\cite{zhang2019rgb}, 
MIDD~\cite{tu2021multi}, 
ECFFNet~\cite{zhou2021ecffnet} and 
SwinNet~\cite{Liu2021swinNet}. 

\begin{table*}[t]
	\centering
	\caption{Performance comparison (F-measure, $ \mathrm{F}_{\beta} \uparrow $) with nine methods on 13 challenges of the VT5000 dataset.
		Bold font highlights the best results in each column.
	}
	\label{tab:chall_res}
	\setlength{\tabcolsep}{3mm}{%
		\begin{tabular}{@{}cccccccccccccc@{}}
			\toprule
			Challenge & BSO            & SSO            & MSO            & CB             & CIB            & OF             & SA             & TC             & IC             & LI             & BW             & RGB            & T              \\ \midrule
			R3Net     & 0.734          & 0.538          & 0.609          & 0.623          & 0.654          & 0.701          & 0.614          & 0.608          & 0.624          & 0.709          & 0.562          & 0.673          & 0.683          \\
			CPD       & 0.835          & 0.694          & 0.765          & 0.777          & 0.799          & 0.801          & 0.756          & 0.789          & 0.764          & 0.823          & 0.694          & 0.804          & 0.805          \\
			PoolNet   & 0.768          & 0.624          & 0.664          & 0.687          & 0.717          & 0.747          & 0.670          & 0.686          & 0.683          & 0.735          & 0.661          & 0.727          & 0.733          \\
			SGDL      & 0.722          & 0.715          & 0.660          & 0.656          & 0.654          & 0.707          & 0.598          & 0.621          & 0.631          & 0.697          & 0.583          & 0.705          & 0.710          \\
			ADF       & 0.858          & 0.737          & 0.806          & 0.821          & 0.837          & 0.806          & 0.791          & 0.792          & 0.803          & 0.845          & 0.771          & 0.840          & 0.842          \\
			FMCF      & 0.815          & 0.559          & 0.724          & 0.740          & 0.782          & 0.743          & 0.701          & 0.723          & 0.725          & 0.745          & 0.698          & 0.762          & 0.763          \\
			MIDD      & 0.848          & 0.696          & 0.781          & 0.803          & 0.818          & 0.799          & 0.755          & 0.778          & 0.768          & 0.797          & 0.756          & 0.817          & 0.817          \\
			ECFFNet   & 0.878          & 0.735          & 0.822          & 0.840          & 0.860          & 0.823          & 0.801          & 0.814          & 0.816          & 0.850          & 0.765          & 0.854          & 0.855          \\
			SwinNet   & 0.919          & 0.839          & 0.882          & 0.895          & 0.910          & 0.890          & 0.884          & 0.886          & 0.875          & 0.914          & 0.863          & 0.903          & 0.906          \\ \midrule
			Ours      & \textbf{0.929} & \textbf{0.874} & \textbf{0.897} & \textbf{0.913} & \textbf{0.924} & \textbf{0.895} & \textbf{0.902} & \textbf{0.908} & \textbf{0.897} & \textbf{0.918} & \textbf{0.881} & \textbf{0.918} & \textbf{0.917} \\ \bottomrule
		\end{tabular}%
	}
\end{table*}

\subsubsection{Qualitative Comparison}
The results visualized in Fig.~\ref{Fig.res} display a qualitative comparison of some challenging image pairs,
such as
SSO (column (a)-(c)), 
CB (column (d) and (e)),
BSO (column (e), (f) and (q)-(t)), 
BW (column (g) and (r)),
TC (column (h)-(j)),
LI (column (f) and (k)), 
MSO (column (k)-(n)), 
SA (column (i)-(l)), 
CIB (column (o)-(r) and (u)), 
IC (column (s) and (v)), 
OF (column (p) and (r)),
RGB images with low quality (column (a), (g), (k), (n) and (v)) 
and 
thermal images with low quality (column (a), (e), (i), (m) and (s)).
As illustrated in Fig.~\ref{Fig.res}, the results of our PRLNet are qualitatively superior to all SOTA methods.
Our method takes full advantage of the discriminative feature representation capabilities of the swin transformer, 
while taking the position relations between pixels into account, \textit{i.e.}, distance and direction relationships.

As shown in Fig.~\ref{Fig.res} (e), (h) and (o), the salient objects and background objects in certain spectral images have similar intensities, which can lead to confusion between foreground and background classes.
Our proposed SDMAM effectively addresses this problem by explicitly constraining the foreground-background difference with signs and modeling the distance of different pixels from the boundary.
SDMAM increases inter-class separability.
From the results in Fig.~\ref{Fig.res} (e), (h) and (o), it can be seen that for background objects similar to the target, such as \textit{cabinets}, \textit{chairs} and \textit{buildings}, our method accurately excludes inter-class interference.

On the other hand, the foreground objects also contain many components with large differences, which leads to inconsistencies in the intra-class features, as shown in Fig.~\ref{Fig.res} (d), (n) and (q).
Our proposed FRDF learns the directional relations of pixels in salient regions, enhancing the intra-class compactness of feature representations.
From the results in Fig.~\ref{Fig.res} (d), (n) and (q), it can be seen that the salient object masks generated by our model are more homogenous compared to other methods.
Overall, as shown in Fig.~\ref{Fig.res} (s), (t), (u) and (v), our PRLNet can generate masks with clear boundaries and smooth regions for objects with fine structures, such as \textit{stone fountains} and \textit{leafy plants}.
The extensive visualization results in Fig.~\ref{Fig.res} effectively prove that our method can handle a variety of complex scenarios with superior performance.
Above all, the saliency masks generated by our PRLNet are consistently the closest to GT.

\subsubsection{Quantitative Comparison}
TABLE~\ref{tab:sota} and Fig.~\ref{Fig.PR} provides a quantitative comparison of our model with other models on three datasets.
First, it can be seen from TABLE~\ref{tab:sota} that our PRLNet achieves the highest results on VT821, VT1000 and VT5000.
This benefits from the fact that our proposed position-aware relation learning can effectively enhance the intra-class compactness and inter-class separability of feature representations.

Specifically, our PRLNet achieves a marked superiority on VT821.
As shown in the results of VT821 in TABLE~\ref{tab:sota}, 
our method improves on average by 0.101, 0.073, 0.152 and 0.043 over other nine methods for 
$ \mathrm{S}_{\alpha} $,
$ \mathrm{F}_{\beta} $,
$ \mathrm{E}_{m} $ and 
$ \mathcal{M} $, respectively.
Compared with other methods on VT1000, PRLNet has an average improvement of 0.063, 0.036, 0.094, and 0.026 on the four metrics, respectively.
As reported in the results of VT5000 from TABLE~\ref{tab:sota}, the performance of our PRLNet has improved by an average of 0.092, 0.067, 0.155, and 0.034 on 
$ \mathrm{S}_{\alpha} $,
$ \mathrm{F}_{\beta} $,
$ \mathrm{E}_{m} $ and 
$ \mathcal{M} $, respectively.
Moreover, for salient masks, 
structural similarity ($ \mathrm{S}_{\alpha} $ and $ \mathrm{E}_{m} $) can better characterize the homogeneity of foreground-background regions and the sharpness of boundaries.
From the above analysis, it can be seen that our PRLNet improves much higher in $ \mathrm{S}_{\alpha} $ and $ \mathrm{E}_{m} $ metrics than other two metrics.
This indicates out that the salient mask of our method is more sophisticated and close to the ground truth.
In addition, compared with the previous state-of-the-art method SwinNet~\cite{zhou2021ecffnet} on three datasets, 
our PRLNet achieves an average gain of 1.02\%, 1.12\%, 0.57\%, 13.11\% w.r.t $ \mathrm{S}_{\alpha} $,
$ \mathrm{F}_{\beta} $,
$ \mathrm{E}_{m} $ and 
$ \mathcal{M} $.

Meanwhile, the P-R curves in Fig.~\ref{Fig.PR} also gives consistent results.
As shown in Fig.~\ref{Fig.PR}, our curves noticeably lie above the others on VT821, VT1000 and VT5000 datasets.
Our proposed method outperforms the state-of-the-art methods.
Above all, both the P-R curves and quantization results on the three datasets demonstrate the validity and advantages of our PRLNet for RGB-T SOD.

\begin{figure*}[t]  
	\centering 
	{\includegraphics[scale=0.31]{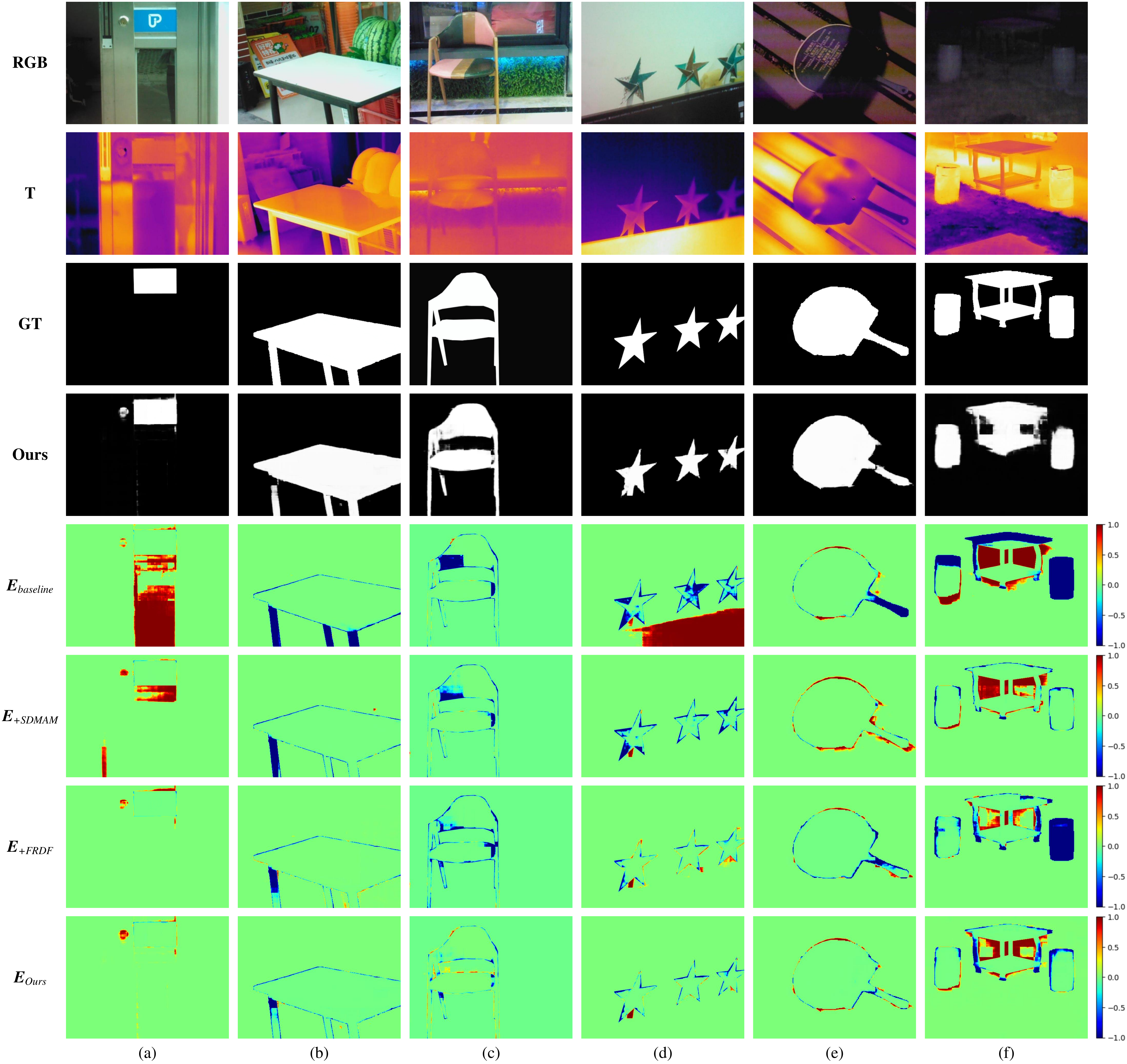}}  
	\caption{
		Visualization results of the error map $ \boldsymbol{E} = \boldsymbol{O}_{sal}-\boldsymbol{G} $,
		where $ \boldsymbol{E}(p) >0 $ indicates a false positive pixel (FP), and $ \boldsymbol{E}(p) < 0 $ indicates a false negative pixel (FN).
		$ \boldsymbol{E}_{\mathrm{\textit{baseline}}} $ represents the error map of the prediction results for the baseline model without SDMAM and FRDF. 
		$ \boldsymbol{E}_{\mathrm{\textit{+SDMAM}}} $ and $ \boldsymbol{E}_{\mathrm{\textit{+FRDF}}} $ represent the error map after using the SDMAM and FRDF modules, respectively.
	}
	\label{Fig.abl} 
\end{figure*}

\subsubsection{Quantitative Comparison on Challenge}
To further validate the performance of our PRLNet, we evaluate the performance of each model on all challenges of VT5000 dataset.
TABLE~\ref{tab:challeng} summarizes the challenges covered by the VT5000.
Challenge-based quantitative comparison results are reported in TABLE~\ref{tab:chall_res}.
The best performance of our PRLNet (Ours) is achieved on all 13 challenges.
Compared with SwinNet, our method achieves an average improvement of 1.81\% on all challenges.
PRLNet achieves an average performance of 0.905 in handling diverse complex targets challenging, such as BSO, SSO, MSO, CB, CIB, OF, SA and TC.

Fig.~\ref{Fig.res} (b), (c), and (q) show the results on challenges on small objects, multi-object, and large object images, respectively.
For example, \textit{soccer goals} and \textit{fences} shown in Fig.~\ref{Fig.res} (q) and (r) are two common BSO challenges.
Compared with the SOTA methods, the BSO mask generated by our PRLNet maintains the global consistency of large objects with better intra-class compactness.
TABLE~\ref{tab:chall_res} reports that our method achieves the highest performance of 0.929 on BSO.
Some of the TC challenges are shown in Fig.~\ref{Fig.res} (h), (i) and (j), targets such as \textit{ragdolls} on wooden chairs, \textit{water bottles} in bicycle baskets, and \textit{cats} on park seats have similar thermal radiation to their surroundings.
The results from Fig.~\ref{Fig.res} suggest that the existing methods have difficulty in detecting \textit{ragdolls}, \textit{water bottles} and \textit{cats} from the background with TC. 
In contrast, our PRLNet obviously suppresses the background objects with thermal crossover, and F-measure attains 0.908 on TC as shown in TABLE~\ref{tab:chall_res}.

The challenges caused by weather or illumination, such as IC, LI and BW, degrade the performance of SOD models.
As can be seen from TABLE~\ref{tab:chall_res}, our model still achieves the best performance of about 0.9 for the degraded scenario.
In addition, for multispectral RGB-T images, PRLNet effectively learns robust cross-spectral fusion features and reduces the interference caused by spectral inconsistency.
The thermal image in Fig.~\ref{Fig.res} (m) and the RGB image in Fig.~\ref{Fig.res} (v) have lower quality than the image in the other spectrum. 
PRLNet overcomes the effect of RGB-T spectral inconsistency and achieves a F-measure above 0.917.

Both the visualization results in Fig.~\ref{Fig.res} and the quantitative comparisons in TABLE~\ref{tab:chall_res} demonstrate that our method can effectively deal with a variety of salient objects.
Above all, the challenge-based quantitative analysis and detailed visualization results consistently demonstrate that 
our method can effectively address various challenges and outperform state-of-the-art methods.

\subsection{Ablation Study}
\label{ssec:ablation}
Our PRLNet mainly contains two key insights: SDM auxiliary module (SDMAM) and feature refinement approach with direction field (FRDF). 
Therefore, we conduct ablation experiments to verify the validity of components and the involved hyper-parameters.

\subsubsection{Effectiveness of SDMAM}
TABLE~\ref{tab:ablation} reports the contributions of different components to the model, 
and Fig.~\ref{Fig.abl} shows the corresponding visualization results.
The first row of TABLE~\ref{tab:ablation} represents the baseline model, which does not use the SDMAM and FRDF modules.
As can be seen from row 2 in TABLE~\ref{tab:ablation}, 
$ \mathrm{S}_{\alpha} $,
$ \mathrm{F}_{\beta} $,
$ \mathrm{E}_{m} $ and 
$ \mathcal{M} $ attain 0.916, 0.868, 0.913 and 0.033, respectively.
SDMAM improves the performance gain by 7.33\% on average across the four metrics on the VT5000 dataset compared to the baseline model.
The boundary discrimination of the features is enhanced by SDMAM to distinguish salient objects from the background.

To further prove the effectiveness and interpretability of our network, we visualize the error maps (\textit{i.e.}, $ \boldsymbol{E}_{\mathrm{\textit{+SDMAM}}} $ and $ \boldsymbol{E}_{\mathrm{\textit{+FRDF}}} $) of the saliency maps generated by different components.
As shown in Fig.~\ref{Fig.abl} (row 6), 
SDMAM visibly reduces the error pixels and strengthens the separability of inter-class features.
The results of $ \boldsymbol{E}_{\mathrm{\textit{+SDMAM}}} $ in the Fig.~\ref{Fig.abl} (a) and (d) illustrate that SDMAM notably suppresses the false alarm (\textit{i.e.}, FP).

\begin{table}[t]
	\centering
	\caption{Ablation study on SDMAM and FRDF on VT5000 dataset.
		Bold font highlights the best results in each column.}
	\label{tab:ablation}
	\setlength{\tabcolsep}{3.5mm}{%
		\begin{tabular}{@{}cccccc@{}}
			\toprule
			$ \mathrm{SDMAM} $ 	&$\mathrm{FRDF} $ 	& {$S_{\alpha} \uparrow$} 	& {$F_{\beta} \uparrow$} 	& {$E_{m} \uparrow$}  	& {$\mathcal{M} \downarrow$}   \\ \midrule
			&      				& 0.904          	& 0.817          & 0.910          & 0.042          \\
			$ \checkmark $     	&      				& 0.916          & 0.868          & 0.913          & 0.033          \\
			& $ \checkmark $    & 0.918          & 0.866          & 0.930          & 0.026          \\
			$ \checkmark $     	& $ \checkmark $    & \textbf{0.921} & \textbf{0.875} & \textbf{0.948} & \textbf{0.023} \\ \bottomrule
		\end{tabular}%
	}
\end{table}

\subsubsection{Effectiveness of FRDF}
As can be seen from row 3 in TABLE~\ref{tab:ablation}, 
$ \mathrm{S}_{\alpha} $,
$ \mathrm{F}_{\beta} $,
$ \mathrm{E}_{m} $ and 
$ \mathcal{M} $ attain 0.918, 0.866, 0.930 and 0.026, respectively.
FRDF brings an average performance gain of 11.96\% over the baseline model.
This suggests that the directional information of object pixels is essential and indispensable for learning a fine feature.
As shown in Fig.~\ref{Fig.abl} (a), (b), and (e), 
the error map of the predicted result with FRDF effectively handles the missed detection (\textit{i.e.}, FN) and generates object masks with clear contour and homogeneous regions.
The visualization results $ \boldsymbol{E}_{\mathrm{\textit{+FRDF}}}$ in Fig.~\ref{Fig.abl} straightforwardly demonstrate the effectiveness of our proposed FRDF.
In Section \ref{sec:method}, we argue that the distance relationship and direction relationship between pixels are crucial for salient object detection, which can be further proved by this experiment.
Above all, we can conclude from TABLE~\ref{tab:ablation} and Fig.~\ref{Fig.abl} that each component is integral and complementary, which together contribute to the final result.

\subsubsection{Hyper-Parameters Analysis}
\label{sssec:hyperpara}

\begin{figure}[t]  
	\centering 
	{\includegraphics[scale=0.9]{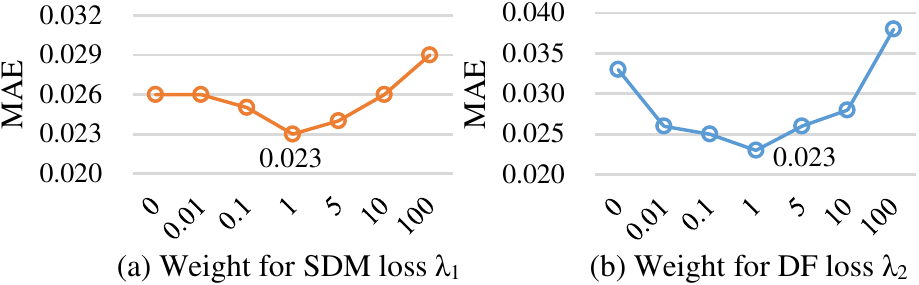}}
	\caption{
		Hyper-parameters analysis of $\lambda_1$ and $\lambda_2$ in the position-aware relation learning loss (PRL loss).
	}
	\label{Fig.loss} 
\end{figure}

The parameters $ \lambda_{1} $ and $ \lambda_{2} $ control the relative importance between SDM loss and DF loss in Eq. \eqref{Eq.loss_all}, which is the decisive hyperparameter for the detection results.
The greater the value, the more importance lies in the proposed PRL loss.
As shown in Fig.~\ref{Fig.loss} (a), MAE score decreases as $ \lambda_{1} $ grows from 0.01 to 1 and increases as $ \lambda_{1} $ grows from 1 to 100, where the valley score reaches 0.023 when $ \lambda_{1}=1 $. 
As $ \lambda_{1} $ becomes larger, the SDM loss dominates the PRL loss, leading to model performance degradation.
The result in Fig.~\ref{Fig.loss} (b) shows that MAE keeps decreasing when $ \lambda_{2} $ increases to 1. 
As $ \lambda_{2} $ grows further, the DF loss dominates model training, leading to an increase in the number of error pixels in the saliency mask.
Hence, we fix $ \lambda_{1}=1 $, $ \lambda_{2} =1$ in the following experimental settings.

\begin{figure}[t]  
	\centering 
	{\includegraphics[scale=0.35]{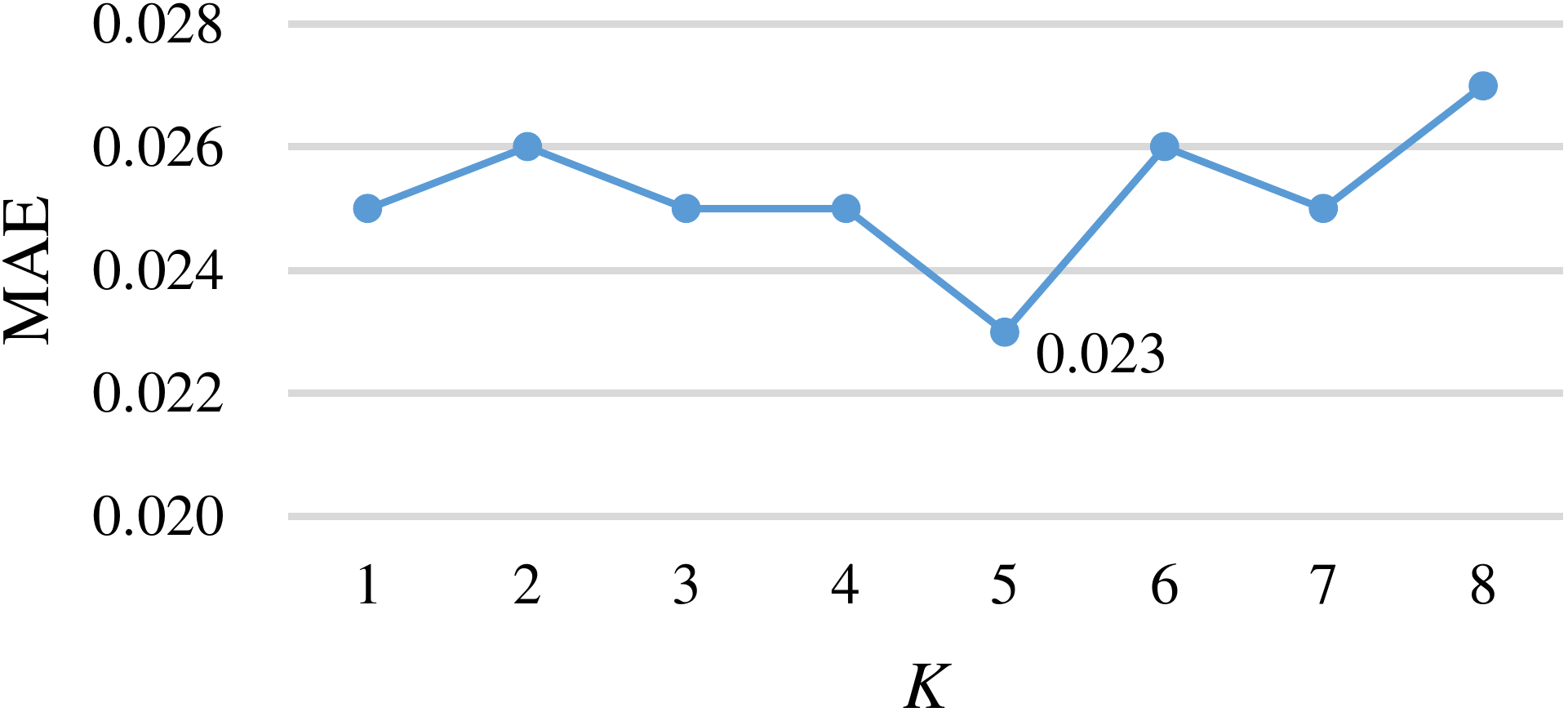}}
	\caption{
		Hyper-parameters analysis of $ K $ in the feature refinement approach with directional field.
	}
	\label{Fig.hp} 
\end{figure}

The number of iterations, \textit{i.e.}, $ K $, is another important hyperparameter in our method.
As $ K $ controls the number of iterations of our proposed FRDF, we conduct experiments to verify the choice of $ K = 5 $ in Eq. \eqref{Eq.zk}.
We vary $ K $ from 0 to 8, as shown in Fig. \ref{Fig.hp}.
The MAE constantly decreases as $ K $ is growing from 1 to 5.
However, there is a slight increase after 5 due to over-refinement with too many iterations.
From the above analysis, we choose $ K = 5 $ as the number of iterations for our feature refinement approach with direction field.

\section{Conclusion}
\label{sec:conclusion}
In this paper, we have proposed novel a position-aware relation learning network (PRLNet) with pure transformer for RGB-T SOD.
PRLNet explored the distance and direction relationships between pixels to strengthen intra-class compactness and inter-class separation.
Specifically, we first constructed a dual-stream encoder and decoder framework based on swin transformer, where a patch separation layer was designed to decode the patches.
Then, we proposed SDMAM to learn the distance relationship between foreground-background regions and boundaries, which enhanced the boundary perception capability of PRLNet.
In addition, we designed FRDF to iteratively rectify the features of the bounding neighborhood using the internal features of the salient objects. FRDF strengthened the intra-class compactness of the salient regions.
Extensive experiments and comparisons have shown that the proposed PRLNet consistently outperforms the state-of-the-art methods on three public RGB-T SOD benchmark datasets.
Notably, visualization results not only demonstrated that the salient masks generated by our PRLNet have sharp boundaries and homogeneous regions, but also offered a new insight to investigate the relationship between pixels.
In future work, we will pay more attention to the following two directions: camouflage object detection (COD) and multispectral image fusion.
Firstly, the proposed relation-aware learning can be applied to COD. In contrast to SOD, COD aims to identify objects embedded in the surrounding environment. Both SOD and COD need to effectively perceive the boundaries of objects and generate masks with clear boundaries and homogeneous regions.
Secondly, we will study the efficient complementary fusion between RGB and thermal images.
The quality of RGB and thermal images under different illumination conditions is modeled by uncertainty.
The poor quality spectral images are used to enhance the good quality spectral images, which enhances the complementarity of multispectral image fusion.

\bibliographystyle{IEEEtran}
\bibliography{ref_zh}

\begin{thebibliography}{10}
\providecommand{\url}[1]{#1}
\csname url@samestyle\endcsname
\providecommand{\newblock}{\relax}
\providecommand{\bibinfo}[2]{#2}
\providecommand{\BIBentrySTDinterwordspacing}{\spaceskip=0pt\relax}
\providecommand{\BIBentryALTinterwordstretchfactor}{4}
\providecommand{\BIBentryALTinterwordspacing}{\spaceskip=\fontdimen2\font plus
\BIBentryALTinterwordstretchfactor\fontdimen3\font minus
  \fontdimen4\font\relax}
\providecommand{\BIBforeignlanguage}[2]{{%
\expandafter\ifx\csname l@#1\endcsname\relax
\typeout{** WARNING: IEEEtran.bst: No hyphenation pattern has been}%
\typeout{** loaded for the language `#1'. Using the pattern for}%
\typeout{** the default language instead.}%
\else
\language=\csname l@#1\endcsname
\fi
#2}}
\providecommand{\BIBdecl}{\relax}
\BIBdecl

\bibitem{gu2016saliency}
K.~Gu, S.~Wang, H.~Yang, W.~Lin, G.~Zhai, X.~Yang, and W.~Zhang,
  ``Saliency-guided quality assessment of screen content images,'' \emph{IEEE
  Transactions on Multimedia}, vol.~18, no.~6, pp. 1098--1110, 2016.

\bibitem{chen2021depth}
C.~Chen, J.~Wei, C.~Peng, and H.~Qin, ``Depth-quality-aware salient object
  detection,'' \emph{IEEE Transactions on Image Processing}, vol.~30, pp.
  2350--2363, 2021.

\bibitem{chen2020rgbd}
H.~Chen, Y.~Deng, Y.~Li, T.-Y. Hung, and G.~Lin, ``Rgbd salient object
  detection via disentangled cross-modal fusion,'' \emph{IEEE Transactions on
  Image Processing}, vol.~29, pp. 8407--8416, 2020.

\bibitem{feng2019attentive}
M.~Feng, H.~Lu, and E.~Ding, ``Attentive feedback network for boundary-aware
  salient object detection,'' in \emph{Proceedings of the IEEE/CVF Conference
  on Computer Vision and Pattern Recognition}, 2019, pp. 1623--1632.

\bibitem{zhou2019discriminative}
S.~Zhou, J.~Wang, D.~Meng, Y.~Liang, Y.~Gong, and N.~Zheng, ``Discriminative
  feature learning with foreground attention for person re-identification,''
  \emph{IEEE Transactions on Image Processing}, vol.~28, no.~9, pp. 4671--4684,
  2019.

\bibitem{dawson2020provably}
C.~Dawson, A.~Jasour, A.~Hofmann, and B.~Williams, ``Provably safe trajectory
  optimization in the presence of uncertain convex obstacles,'' in \emph{2020
  IEEE/RSJ International Conference on Intelligent Robots and Systems
  (IROS)}.\hskip 1em plus 0.5em minus 0.4em\relax IEEE, 2020, pp. 6237--6244.

\bibitem{zhou2020hierarchical}
S.~Zhou, J.~Wang, L.~Wang, J.~Zhang, F.~Wang, D.~Huang, and N.~Zheng,
  ``Hierarchical and interactive refinement network for edge-preserving salient
  object detection,'' \emph{IEEE Transactions on Image Processing}, vol.~30,
  pp. 1--14, 2020.

\bibitem{wang2021salient}
W.~Wang, Q.~Lai, H.~Fu, J.~Shen, H.~Ling, and R.~Yang, ``Salient object
  detection in the deep learning era: An in-depth survey,'' \emph{IEEE
  Transactions on Pattern Analysis and Machine Intelligence}, vol.~44, no.~6,
  pp. 3239--3259, 2021.

\bibitem{zhang2019rgb}
Q.~Zhang, N.~Huang, L.~Yao, D.~Zhang, C.~Shan, and J.~Han, ``Rgb-t salient
  object detection via fusing multi-level cnn features,'' \emph{IEEE
  Transactions on Image Processing}, vol.~29, pp. 3321--3335, 2019.

\bibitem{liu2020multi}
Q.~Liu, X.~Li, Z.~He, N.~Fan, D.~Yuan, W.~Liu, and Y.~Liang, ``Multi-task
  driven feature models for thermal infrared tracking,'' in \emph{Proceedings
  of the AAAI Conference on Artificial Intelligence}, vol.~34, no.~07, 2020,
  pp. 11\,604--11\,611.

\bibitem{zhang2021summarize}
N.~Zhang, J.~Han, N.~Liu, and L.~Shao, ``Summarize and search: Learning
  consensus-aware dynamic convolution for co-saliency detection,'' in
  \emph{Proceedings of the IEEE/CVF International Conference on Computer
  Vision}, 2021, pp. 4167--4176.

\bibitem{zhou2022edge}
W.~Zhou, S.~Dong, C.~Xu, and Q.~Yaguan, ``Edge-aware guidance fusion network
  for rgb–thermal scene parsing,'' in \emph{Proceedings of the AAAI
  Conference on Artificial Intelligence}, 2022.

\bibitem{song2020multi}
S.~Song, H.~Yu, Z.~Miao, J.~Fang, K.~Zheng, C.~Ma, and S.~Wang,
  ``Multi-spectral salient object detection by adversarial domain adaptation,''
  in \emph{Proceedings of the AAAI Conference on Artificial Intelligence},
  vol.~34, no.~07, 2020, pp. 12\,023--12\,030.

\bibitem{bondi2020birdsai}
E.~Bondi, R.~Jain, P.~Aggrawal, S.~Anand, R.~Hannaford, A.~Kapoor, J.~Piavis,
  S.~Shah, L.~Joppa, B.~Dilkina \emph{et~al.}, ``Birdsai: A dataset for
  detection and tracking in aerial thermal infrared videos,'' in
  \emph{Proceedings of the IEEE/CVF Winter Conference on Applications of
  Computer Vision}, 2020, pp. 1747--1756.

\bibitem{hao2019hsme}
Y.~Hao, N.~Wang, J.~Li, and X.~Gao, ``Hsme: Hypersphere manifold embedding for
  visible thermal person re-identification,'' in \emph{Proceedings of the AAAI
  Conference on Artificial Intelligence}, vol.~33, no.~01, 2019, pp.
  8385--8392.

\bibitem{zhou2022multi}
H.~Zhou, C.~Tian, Z.~Zhang, Q.~Huo, Y.~Xie, and Z.~Li, ``Multi-spectral fusion
  transformer network for rgb-thermal urban scene semantic segmentation,''
  \emph{IEEE Geoscience and Remote Sensing Letters}, vol.~19, pp. 1--5, 2022.

\bibitem{qin2019basnet}
X.~Qin, Z.~Zhang, C.~Huang, C.~Gao, M.~Dehghan, and M.~Jagersand, ``Basnet:
  Boundary-aware salient object detection,'' in \emph{Proceedings of the
  IEEE/CVF Conference on Computer Vision and Pattern Recognition}, 2019, pp.
  7479--7489.

\bibitem{zhang2022collaborative}
Z.~Zhang, C.~Tian, X.~Gao, J.~Li, Z.~Jiao, C.~Wang, and Z.~Zhong,
  ``Collaborative boundary-aware context encoding networks for error map
  prediction,'' \emph{Pattern Recognition}, vol. 125, p. 108515, 2022.

\bibitem{zhang2021location}
X.~Zhang, B.~Ma, H.~Chang, S.~Shan, and X.~Chen, ``Location sensitive network
  for human instance segmentation,'' \emph{IEEE Transactions on Image
  Processing}, vol.~30, pp. 7649--7662, 2021.

\bibitem{zhang2019resls}
W.~Zhang, X.~Wang, W.~You, J.~Chen, P.~Dai, and P.~Zhang, ``Resls: Region and
  edge synergetic level set framework for image segmentation,'' \emph{IEEE
  Transactions on Image Processing}, vol.~29, pp. 57--71, 2019.

\bibitem{cai2021avlsm}
Q.~Cai, Y.~Qian, S.~Zhou, J.~Li, Y.-H. Yang, F.~Wu, and D.~Zhang, ``Avlsm:
  Adaptive variational level set model for image segmentation in the presence
  of severe intensity inhomogeneity and high noise,'' \emph{IEEE Transactions
  on Image Processing}, vol.~31, pp. 43--57, 2021.

\bibitem{farag2013novel}
A.~A. Farag, H.~E. Abd El~Munim, J.~H. Graham, and A.~A. Farag, ``A novel
  approach for lung nodules segmentation in chest ct using level sets,''
  \emph{IEEE Transactions on Image Processing}, vol.~22, no.~12, pp.
  5202--5213, 2013.

\bibitem{chai2020aerial}
D.~Chai, S.~Newsam, and J.~Huang, ``Aerial image semantic segmentation using
  dcnn predicted distance maps,'' \emph{ISPRS Journal of Photogrammetry and
  Remote Sensing}, vol. 161, pp. 309--322, 2020.

\bibitem{lin2021bsda}
L.~Lin, Z.~Wang, J.~Wu, Y.~Huang, J.~Lyu, P.~Cheng, J.~Wu, and X.~Tang,
  ``Bsda-net: A boundary shape and distance aware joint learning framework for
  segmenting and classifying octa images,'' in \emph{International Conference
  on Medical Image Computing and Computer-Assisted Intervention}.\hskip 1em
  plus 0.5em minus 0.4em\relax Springer, 2021, pp. 65--75.

\bibitem{cheng2020learning}
F.~Cheng, C.~Chen, Y.~Wang, H.~Shi, Y.~Cao, D.~Tu, C.~Zhang, and Y.~Xu,
  ``Learning directional feature maps for cardiac mri segmentation,'' in
  \emph{International Conference on Medical Image Computing and
  Computer-Assisted Intervention}.\hskip 1em plus 0.5em minus 0.4em\relax
  Springer, 2020, pp. 108--117.

\bibitem{godard2017unsupervised}
C.~Godard, O.~Mac~Aodha, and G.~J. Brostow, ``Unsupervised monocular depth
  estimation with left-right consistency,'' in \emph{Proceedings of the IEEE
  Conference on Computer Vision and Pattern Recognition}, 2017, pp. 270--279.

\bibitem{wang2018occlusion}
Y.~Wang, Y.~Yang, Z.~Yang, L.~Zhao, P.~Wang, and W.~Xu, ``Occlusion aware
  unsupervised learning of optical flow,'' in \emph{Proceedings of the IEEE
  Conference on Computer Vision and Pattern Recognition}, 2018, pp. 4884--4893.

\bibitem{vaswani2017attention}
A.~Vaswani, N.~Shazeer, N.~Parmar, J.~Uszkoreit, L.~Jones, A.~N. Gomez,
  {\L}.~Kaiser, and I.~Polosukhin, ``Attention is all you need,''
  \emph{Advances in Neural Information Processing Systems}, vol.~30, 2017.

\bibitem{liu2021swin}
Z.~Liu, Y.~Lin, Y.~Cao, H.~Hu, Y.~Wei, Z.~Zhang, S.~Lin, and B.~Guo, ``Swin
  transformer: Hierarchical vision transformer using shifted windows,'' in
  \emph{Proceedings of the IEEE/CVF International Conference on Computer
  Vision}, 2021, pp. 10\,012--10\,022.

\bibitem{he2022swin}
X.~He, Y.~Zhou, J.~Zhao, D.~Zhang, R.~Yao, and Y.~Xue, ``Swin transformer
  embedding unet for remote sensing image semantic segmentation,'' \emph{IEEE
  Transactions on Geoscience and Remote Sensing}, vol.~60, pp. 1--15, 2022.

\bibitem{yu2022soit}
X.~Yu, D.~Shi, X.~Wei, Y.~Ren, T.~Ye, and W.~Tan, ``Soit: Segmenting objects
  with instance-aware transformers,'' in \emph{Proceedings of the AAAI
  Conference on Artificial Intelligence}, vol.~36, no.~3, 2022, pp. 3188--3196.

\bibitem{long2015fully}
J.~Long, E.~Shelhamer, and T.~Darrell, ``Fully convolutional networks for
  semantic segmentation,'' in \emph{Proceedings of the IEEE Conference on
  Computer Vision and Pattern Recognition}, 2015, pp. 3431--3440.

\bibitem{chen2020reverse}
S.~Chen, X.~Tan, B.~Wang, H.~Lu, X.~Hu, and Y.~Fu, ``Reverse attention-based
  residual network for salient object detection,'' \emph{IEEE Transactions on
  Image Processing}, vol.~29, pp. 3763--3776, 2020.

\bibitem{zhang2018deep}
L.~Zhang, X.~Fang, H.~Bo, T.~Wang, and H.~Lu, ``Deep multi-level networks with
  multi-task learning for saliency detection,'' \emph{Neurocomputing}, vol.
  312, pp. 229--238, 2018.

\bibitem{deng2018r3net}
Z.~Deng, X.~Hu, L.~Zhu, X.~Xu, J.~Qin, G.~Han, and P.-A. Heng, ``R3net:
  Recurrent residual refinement network for saliency detection,'' in
  \emph{Proceedings of the 27th International Joint Conference on Artificial
  Intelligence}.\hskip 1em plus 0.5em minus 0.4em\relax AAAI Press Menlo Park,
  CA, USA, 2018, pp. 684--690.

\bibitem{wu2022edn}
Y.-H. Wu, Y.~Liu, L.~Zhang, M.-M. Cheng, and B.~Ren, ``Edn: Salient object
  detection via extremely-downsampled network,'' \emph{IEEE Transactions on
  Image Processing}, vol.~31, pp. 3125--3136, 2022.

\bibitem{li2016deepsaliency}
X.~Li, L.~Zhao, L.~Wei, M.-H. Yang, F.~Wu, Y.~Zhuang, H.~Ling, and J.~Wang,
  ``Deepsaliency: Multi-task deep neural network model for salient object
  detection,'' \emph{IEEE Transactions on Image Processing}, vol.~25, no.~8,
  pp. 3919--3930, 2016.

\bibitem{wu2019cascaded}
Z.~Wu, L.~Su, and Q.~Huang, ``Cascaded partial decoder for fast and accurate
  salient object detection,'' in \emph{Proceedings of the IEEE/CVF Conference
  on Computer Vision and Pattern Recognition}, 2019, pp. 3907--3916.

\bibitem{liu2019simple}
J.-J. Liu, Q.~Hou, M.-M. Cheng, J.~Feng, and J.~Jiang, ``A simple pooling-based
  design for real-time salient object detection,'' in \emph{Proceedings of the
  IEEE/CVF Conference on Computer Vision and Pattern Recognition}, 2019, pp.
  3917--3926.

\bibitem{wei2021syncretic}
Z.~Wei, X.~Yang, N.~Wang, and X.~Gao, ``Syncretic modality collaborative
  learning for visible infrared person re-identification,'' in
  \emph{Proceedings of the IEEE/CVF International Conference on Computer
  Vision}, 2021, pp. 225--234.

\bibitem{wang2018rgb}
G.~Wang, C.~Li, Y.~Ma, A.~Zheng, J.~Tang, and B.~Luo, ``Rgb-t saliency
  detection benchmark: Dataset, baselines, analysis and a novel approach,'' in
  \emph{Chinese Conference on Image and Graphics Technologies}.\hskip 1em plus
  0.5em minus 0.4em\relax Springer, 2018, pp. 359--369.

\bibitem{tu2019m3s}
Z.~Tu, T.~Xia, C.~Li, Y.~Lu, and J.~Tang, ``M3s-nir: Multi-modal multi-scale
  noise-insensitive ranking for rgb-t saliency detection,'' in \emph{2019 IEEE
  Conference on Multimedia Information Processing and Retrieval (MIPR)}.\hskip
  1em plus 0.5em minus 0.4em\relax IEEE, 2019, pp. 141--146.

\bibitem{gao2021unified}
W.~Gao, G.~Liao, S.~Ma, G.~Li, Y.~Liang, and W.~Lin, ``Unified information
  fusion network for multi-modal rgb-d and rgb-t salient object detection,''
  \emph{IEEE Transactions on Circuits and Systems for Video Technology},
  vol.~32, no.~4, pp. 2091--2106, 2021.

\bibitem{tu2022rgbt}
Z.~Tu, Y.~Ma, Z.~Li, C.~Li, J.~Xu, and Y.~Liu, ``Rgbt salient object detection:
  A large-scale dataset and benchmark,'' \emph{IEEE Transactions on
  Multimedia}, 2022.

\bibitem{wang2021cgfnet}
J.~Wang, K.~Song, Y.~Bao, L.~Huang, and Y.~Yan, ``Cgfnet: Cross-guided fusion
  network for rgb-t salient object detection,'' \emph{IEEE Transactions on
  Circuits and Systems for Video Technology}, vol.~32, no.~5, pp. 2949--2961,
  2021.

\bibitem{huo2021efficient}
F.~Huo, X.~Zhu, L.~Zhang, Q.~Liu, and Y.~Shu, ``Efficient context-guided
  stacked refinement network for rgb-t salient object detection,'' \emph{IEEE
  Transactions on Circuits and Systems for Video Technology}, vol.~32, no.~5,
  pp. 3111--3124, 2021.

\bibitem{liang2022multi}
Y.~Liang, G.~Qin, M.~Sun, J.~Qin, J.~Yan, and Z.~Zhang, ``Multi-modal
  interactive attention and dual progressive decoding network for rgb-d/t
  salient object detection,'' \emph{Neurocomputing}, vol. 490, pp. 132--145,
  2022.

\bibitem{tu2019rgb}
Z.~Tu, T.~Xia, C.~Li, X.~Wang, Y.~Ma, and J.~Tang, ``Rgb-t image saliency
  detection via collaborative graph learning,'' \emph{IEEE Transactions on
  Multimedia}, vol.~22, no.~1, pp. 160--173, 2019.

\bibitem{tu2021multi}
Z.~Tu, Z.~Li, C.~Li, Y.~Lang, and J.~Tang, ``Multi-interactive dual-decoder for
  rgb-thermal salient object detection,'' \emph{IEEE Transactions on Image
  Processing}, vol.~30, pp. 5678--5691, 2021.

\bibitem{zhou2021ecffnet}
W.~Zhou, Q.~Guo, J.~Lei, L.~Yu, and J.-N. Hwang, ``Ecffnet: Effective and
  consistent feature fusion network for rgb-t salient object detection,''
  \emph{IEEE Transactions on Circuits and Systems for Video Technology},
  vol.~32, no.~3, pp. 1224--1235, 2021.

\bibitem{zhang2022learning}
N.~Zhang, J.~Han, and N.~Liu, ``Learning implicit class knowledge for rgb-d
  co-salient object detection with transformers,'' \emph{IEEE Transactions on
  Image Processing}, 2022.

\bibitem{dosovitskiy2021an}
\BIBentryALTinterwordspacing
A.~Dosovitskiy, L.~Beyer, A.~Kolesnikov, D.~Weissenborn, X.~Zhai,
  T.~Unterthiner, M.~Dehghani, M.~Minderer, G.~Heigold, S.~Gelly, J.~Uszkoreit,
  and N.~Houlsby, ``An image is worth 16x16 words: Transformers for image
  recognition at scale,'' in \emph{International Conference on Learning
  Representations}, 2021. [Online]. Available:
  \url{https://openreview.net/forum?id=YicbFdNTTy}
\BIBentrySTDinterwordspacing

\bibitem{carion2020end}
N.~Carion, F.~Massa, G.~Synnaeve, N.~Usunier, A.~Kirillov, and S.~Zagoruyko,
  ``End-to-end object detection with transformers,'' in \emph{European
  Conference on Computer Vision}.\hskip 1em plus 0.5em minus 0.4em\relax
  Springer, 2020, pp. 213--229.

\bibitem{zeng2022dual}
C.~Zeng and S.~Kwong, ``Dual swin-transformer based mutual interactive network
  for rgb-d salient object detection,'' \emph{arXiv preprint arXiv:2206.03105},
  2022.

\bibitem{Liu2021swinNet}
Z.~Liu, Y.~Tan, Q.~He, and Y.~Xiao, ``Swinnet: Swin transformer drives
  edge-aware rgb-d and rgb-t salient object detection,'' \emph{IEEE
  Transactions on Circuits and Systems for Video Technology}, vol.~32, no.~7,
  pp. 4486--4497, 2022.

\bibitem{zhu2022dftr}
H.~Zhu, X.~Sun, Y.~Li, K.~Ma, S.~K. Zhou, and Y.~Zheng, ``Dftr:
  Depth-supervised hierarchical feature fusion transformer for salient object
  detection,'' \emph{arXiv preprint arXiv:2203.06429}, 2022.

\bibitem{liu2019deep}
Y.~Liu, J.~Han, Q.~Zhang, and C.~Shan, ``Deep salient object detection with
  contextual information guidance,'' \emph{IEEE Transactions on Image
  Processing}, vol.~29, pp. 360--374, 2019.

\bibitem{lin2013network}
M.~Lin, Q.~Chen, and S.~Yan, ``Network in network,'' \emph{arXiv preprint
  arXiv:1312.4400}, 2013.

\bibitem{zhang2022discriminative}
Z.~Zhang, C.~Tian, H.~X. Bai, Z.~Jiao, and X.~Tian, ``Discriminative error
  prediction network for semi-supervised colon gland segmentation,''
  \emph{Medical Image Analysis}, vol.~79, p. 102458, 2022.

\bibitem{zhang2020dense}
Q.~Zhang, R.~Cong, C.~Li, M.-M. Cheng, Y.~Fang, X.~Cao, Y.~Zhao, and S.~Kwong,
  ``Dense attention fluid network for salient object detection in optical
  remote sensing images,'' \emph{IEEE Transactions on Image Processing},
  vol.~30, pp. 1305--1317, 2020.

\bibitem{fan2017structure}
D.-P. Fan, M.-M. Cheng, Y.~Liu, T.~Li, and A.~Borji, ``Structure-measure: A new
  way to evaluate foreground maps,'' in \emph{Proceedings of the IEEE
  International Conference on Computer Vision}, 2017, pp. 4548--4557.

\bibitem{achanta2009frequency}
R.~Achanta, S.~Hemami, F.~Estrada, and S.~Susstrunk, ``Frequency-tuned salient
  region detection,'' in \emph{2009 IEEE Conference on Computer Vision and
  Pattern Recognition}, 2009, pp. 1597--1604.

\bibitem{fan2018enhanced}
D.-P. Fan, C.~Gong, Y.~Cao, B.~Ren, M.-M. Cheng, and A.~Borji,
  ``Enhanced-alignment measure for binary foreground map evaluation,'' in
  \emph{Proceedings of the Twenty-Seventh International Joint Conference on
  Artificial Intelligence}, 2018, pp. 698--704.

\bibitem{perazzi2012saliency}
F.~Perazzi, P.~Kr{\"a}henb{\"u}hl, Y.~Pritch, and A.~Hornung, ``Saliency
  filters: Contrast based filtering for salient region detection,'' in
  \emph{Proceedings of the IEEE Conference on Computer Vision and Pattern
  Recognition}, 2012, pp. 733--740.

\end{thebibliography}

\vfill

\end{document}